\documentclass[11pt]{article}
\usepackage[margin=1in]{geometry}
\usepackage{hyperref}
\usepackage{times}

\usepackage{bm}
\usepackage{mathtools}
\usepackage{xfrac}
\usepackage{bbm}
\usepackage{cleveref}
\usepackage{tikz}
\usepackage{breqn}

\renewcommand{\vec}{\bm}

\newcommand{\placeholderCount}{m}
\newcommand{\errorTolerance}{\epsilon}

\newcommand{\errorConfidence}{\delta}

\DeclarePairedDelimiterX{\infdivx}[2]{(}{)}{%
  #1\;\delimsize\|\;#2
}

\newcommand{\sampleElement}{x}
\newcommand{\labelElement}{y}
\newcommand{\prediction}{\hat{y}}
\newcommand{\weightsElement}{w}
\newcommand{\weight}{\weightsElement}

\newcommand{\bigOh}{O}

\newcommand{\loss}{\ell}

\newcommand{\sampleSpace}{\mathcal{X}}
\newcommand{\labelSpace}{\mathcal{Y}}
\newcommand{\sampleDist}{\mathcal{D}}
\newcommand{\drawnFrom}{\sim}

\newcommand{\multicalibration}{\mathrm{mc}}

\newcommand{\calibrationError}{\mathrm{c}}
\newcommand{\empiricalCalibrationError}{\hat{\mathrm{c}}}
\newcommand{\Frobenius}{\mathrm{F}}

\newcommand{\scalarBound}{B}

\newcommand{\sampleComplexity}{m}

\newcommand{\rademacherRV}{\sigma}

\newcommand{\sgn}{\mathrm{sgn}}

\newcommand{\E}{\mathop{\mathbb{E}}}
\renewcommand{\Pr}{\mathop{\mathbb{P}}}

\newcommand{\intersection}{\cap}

\newcommand{\weightMatrix}{\vec{W}}

\newcommand{\given}{\mid}

\newcommand{\rademacherComplexity}{\mathcal{R}}

\newcommand{\depth}{T}

\newcommand{\indicatorFunction}{\mathbbm{1}}

\newcommand{\kernel}{\kernelFunction}

\crefname{fact}{fact}{facts}

\DeclarePairedDelimiter{\set}{\{}{\}}

\DeclarePairedDelimiter{\abs}{\lvert}{\rvert}

\DeclarePairedDelimiter{\paren}{(}{)}

\newcommand{\sample}{\vec{\sampleElement}}

\newcommand{\bigTheta}{\Theta}

\newcommand{\weights}{\vec{\weightsElement}}

\newcommand{\realNumbers}{\mathbb{R}}
\newcommand{\naturalNumbers}{\mathbb{N}}

\newcommand{\dimension}{d}
\newcommand{\trainingData}{S}
\newcommand{\dataset}{S}

\newcommand{\train}{\mathrm{train}}
\newcommand{\test}{\mathrm{test}}

\newcommand{\Representativeness}{\mathrm{Rep}}

\newcommand{\marker}{\dagger}

\newcommand{\numSamples}{N}

\newcommand{\placeholderSet}{\mathcal{Z}}

\newcommand{\placeholder}{z}

\newcommand{\risk}{L}
\newcommand{\hyp}{\mathrm{h}}

\newcommand{\hypothesisClass}{\mathcal{H}}
\newcommand{\definedAs}{\triangleq}

\newcommand{\kernelFunction}{K}
\newcommand{\kernelMapping}{\Phi}
\newcommand{\norm}[1]{\left\lVert#1\right\rVert}
\newcommand{\union}{\cup}

\newcommand{\graphDimension}{\mathrm{p}}

\newcommand{\predictionValue}{\prediction}

\newcommand{\group}{\mathsf{g}}
\newcommand{\groups}{\mathsf{G}}

\newcommand{\transpose}{\mathsf{T}}

\newcommand{\neuralNetwork}{\mathrm{NN}}
\newcommand{\nonlinearity}{\sigma}

\newcommand\MTkillspecial[1]{% helper macro
\bgroup
\catcode`\&=9
\let\\\relax%
\scantokens{#1}%
\egroup
}
\DeclarePairedDelimiter\brparen
\lparen\rparen
\reDeclarePairedDelimiterInnerWrapper\brparen{star}{
\mathopen{#1\vphantom{\MTkillspecial{#2}}\kern-\nulldelimiterspace\right.}
#2
\mathclose{\left.\kern-\nulldelimiterspace\vphantom{\MTkillspecial{#2}}#3}}

\usepackage{amsmath,amsfonts,bm}

\def\eqref#1{equation~\ref{#1}}
\def\1{\bm{1}}

\DeclareMathAlphabet{\mathsfit}{\encodingdefault}{\sfdefault}{m}{sl}
\SetMathAlphabet{\mathsfit}{bold}{\encodingdefault}{\sfdefault}{bx}{n}

\usepackage{amsthm}
\usepackage{amsmath}
\usepackage{amssymb}

\theoremstyle{plain}% default
\newtheorem{theorem}{Theorem}
\newtheorem{lemma}[theorem]{Lemma}
\theoremstyle{definition}
\newtheorem{definition}{Definition}
\usepackage{subcaption}
\graphicspath{ {./figures/} }
\begin{document}
\date{}
\title{An Exploration of Multicalibration Uniform Convergence Bounds}
\author{
  Harrison Rosenberg\\
    University of Wisconsin--Madison\\
    \texttt{hrosenberg@ece.wisc.edu}
  \and
  Robi Bhattacharjee\\
  University of California, San Diego \\
  \texttt{rcbhatta@eng.ucsd.edu} 
  \and
  Kassem Fawaz \\
  University of Wisconsin -- Madison \\
  \texttt{kfawaz@wisc.edu}
  \and
  Somesh Jha \\ 
  University of Wisconsin -- Madison \\
  \texttt{jha@cs.wisc.edu}
}
\maketitle
\begin{abstract}
{\color{black} Recent works have investigated the sample complexity necessary for
fair machine learning.  The most advanced of such sample complexity
bounds are developed by analyzing multicalibration uniform convergence
for a given predictor class.  We present a framework which yields
multicalibration error uniform convergence bounds by reparametrizing sample complexities for Empirical Risk Minimization (ERM) learning. From this framework, we demonstrate that
multicalibration error exhibits dependence on the classifier architecture as well as
the underlying data distribution.  We perform an experimental
evaluation to investigate the behavior of multicalibration error for
different families of classifiers.  We compare the results of this
evaluation to multicalibration error concentration bounds.  Our
investigation provides additional perspective on both algorithmic
fairness and multicalibration error convergence bounds.  Given the prevalence of ERM sample complexity bounds, our proposed framework enables machine learning practitioners to easily understand the convergence behavior of multicalibration error for a myriad of classifier architectures.
}\end{abstract}

% !TeX root = ../note.tex

\section{Introduction}

As machine learning (ML) systems have exhibited exemplary performance across a wide variety of tasks, the scope of their use has broadened to include decision-making in healthcare, education, financial, and legal settings. This intertwining of society and machine learning has surfaced the issue of fairness of such ML systems. ML fairness studies whether a given predictor achieves performance parity with respect to a sensitive attribute at the individual or group level~\cite{ind_group_fairness}. In this paper, we focus on characterizing fairness at the group level. Our focus is on understanding the performance of a predictor on different demographic groups.

Among the most common characterizations of fairness are sample complexities for multicalibration error convergence. Multicalibration error is a notion of group fairness; it refers to the discrepancy, for a given population group, between the predicted label and the average of realized predicted labels of all samples within the group. Informally speaking, sample complexities are functions that return the number of samples necessary to achieve a specified performance guarantee associated with Empirical Risk Minimization (ERM). Sample complexity offers a probabilistic certificate on the difference between a predictor's true and empirical risk. Here, we study sample complexities for multicalibration error convergence: how many samples are needed to achieve an equitable performance of machine learning predictors over specified population groups.

Multicalibration error is a comprehensive framework that captures group fairness. Moreover, multicalibration error can be decoupled from specific training constraints and prediction accuracy. Since Hebert-Johnson et al. \cite{hebert2017calibration} introduced the notion of multicalibration error in ML, there has been much follow-up investigation about its relationship with fairness \cite{shabat2020sample,pmlr-v134-jung21a,liu2019implicit}. Thus far, multicalibration error sample complexity bounds require intensive mathematical analysis. Such analysis must take place for each notion of predictor class complexity, such as VC-dimension or Rademacher complexity. In some cases, this same analysis must be repeated for each predictor class; for example, sample complexity bounds for a Linear Support Vector Machine (SVM) may require a separate derivation from that of a two-layer Rectified Linear Unit (ReLU) network.

This paper shows how to achieve sample complexity bounds for multicalibration uniform convergence through reparamterizing sample complexities for ERM learning, which frequently appear in the literature. We prove an explicit relationship between ERM learning sample complexity bounds and bounds for multicalibration error uniform convergence. This explicit relationship, \cref{thm:maintheorem}, grants ML practitioners a plug-and-play technique to capture the multicalibration error convergence behavior. For example, when applying our main theorem to VC dimension-based ERM bounds, we arrive at the multicalibration uniform convergence bounds by Shabat et al.~\cite{shabat2020sample}. Further, we apply our theorem to ERM sample complexity bounds based on Rademacher complexity for SVMs and two-layer ReLU networks.

Our main result implies a dependence on both dataset composition and choice of classifier architecture. Consequently, a simple classifier such as a linear SVM takes fewer samples to converge in multicalibration error than a more complicated model such as a deep neural network. Likewise, our results suggest dataset composition influences multicalibration error convergence. The least frequent group controls the sample complexity necessary to see multicalibration uniform convergence; balanced datasets lead to faster convergence. We experimentally validate our findings on tabular and image datasets. Our results in \cref{sec:evaluation} demonstrate, for each category, that when the number of samples increases, the empirical calibration error converges towards the true calibration error. Interestingly, we observe that the convergence behavior of the multicalibration error differs from that indicated by sample complexity bounds; our results suggest that calibration error may converge faster on neural network classifiers than on SVMs, which is counter to our theory. We attribute this discrepancy to ERM bounds which are known to be loose for ReLU networks.

 To summarize, we put forth a framework for ML practitioners to construct multicalibration uniform convergence bounds from sample complexities for ERM learning. As we will show, these bounds have an explicit dependence on the frequency of the least frequent protected group. This paper makes the following contributions:

\begin{enumerate}
    
    \item We show that sample complexities for state-of-the-art multicalibration uniform convergence are essentially re-parametrizations of sample complexities for ERM learning of hypothesis classes. As an example, we instantiate, via the Rademacher complexity, multicalibration bounds on small neural networks and Radial Basis Function (RBF) kernel SVMs (\cref{sec:our_bound}).
    
    \item We draw insights into dependencies of multicalibration error. We find that structural issues in both the dataset and the choice of classifier architecture induce a baseline of unfairness that cannot be overcome with any training regime.

    \item We perform empirical evaluation on several representative datasets: Adult from UCI \cite{Dua:2019}, COMPAS from ProPublica \cite{dressel2018accuracy,dieterich2016compas}, and CelebA from Chinese University of Hong Kong \cite{liu2015faceattributes}
    % \footnote{Our codebase will be made public upon paper publication}
    . We study the tightness of the presented bounds and draw insights about the impact of both the data distribution and classifier regime on fairness. We show that bounds for multicalibration error convergence can be loose quite loose, especially for bounds on neural network classes.

\end{enumerate}

\section{Background}\label{sec:background}

Let us put forth the notation with which we will analyze multicalibration error: $\sampleSpace \subseteq \realNumbers^{\dimension}$ denotes the space of examples and $\labelSpace$ denotes the set of possible labels.  A sequence of examples is denoted as $\trainingData{} \subseteq \sampleSpace \times \labelSpace{}$ and contains the $\numSamples$ examples which constitute the population group.  $\trainingData$ is  $\left\{(\sample_1,\labelElement_1),\dotsc,(\sample_\numSamples,\labelElement_\numSamples)\right\}$.  Sometimes we abuse notation and use $\trainingData$ to represent an unlabeled set of training examples, i.e. $\trainingData \definedAs \left\{\sample_1,\dotsc,\sample_\numSamples \right\}$.  We may also refer to $\trainingData$ as a dataset.

The classifier architecture defines a predictor class $\hypothesisClass$, from which a predictor $\hyp : \sampleSpace \to \labelSpace$ is selected.  The terms hypothesis and predictor are used interchangably. This predictor selection process is known as \emph{training} and is performed by ERM, also known as loss minimization.  A loss function $\loss: \labelSpace \times \labelSpace \to \realNumbers$ is often used as a surrogate for accuracy.  Common examples of loss functions include logistic loss, hinge loss, and cross-entropy loss functions.  The \emph{risk} refers to the expected loss of a predictor $\hyp$.  More specifically, there two related varieties of risk to which we refer: the \emph{empirical risk} and the \emph{true risk}.  The empirical risk $\risk_{\trainingData}(\hyp)$ refers to the expected loss of a predictor $\hyp$ taken over a finite sample $\trainingData$ drawn from a data distribution $\sampleDist$:

\begin{equation}\label{eq:empiricalRisk}
    \risk_{\trainingData}(\hyp) \definedAs \frac{1}{\numSamples}\sum_{(\sample,\labelElement)\in \trainingData} \loss\left(\hyp(\sample),\labelElement\right)
\end{equation}

The true risk $\risk_{\sampleDist}(\hyp)$ refers to the expected loss of a predictor taken over a typically unknown data distribution $\sampleDist$:

\begin{equation}\label{eq:trueRisk}
    \risk_{\sampleDist}(\hyp) \definedAs \E_{(\sample,\labelElement) \drawnFrom \trainingData} \loss\left(\hyp(\sample),\labelElement\right)
\end{equation}

Because $\trainingData$ is drawn i.i.d. from $\sampleDist$, $\risk_{\trainingData}(\hyp)$ is a consistent estimator of $ \risk_{\sampleDist}(\hyp)$.

The overall population $\trainingData$ breaks into potentially overlapping population groups denoted as $\group_i$.  The set of population groups is denoted $\groups$ and can be decomposed as $\set{\group_1,\dotsc,\group_{\abs{\groups}}}$. Each group $\group_i$ is associated with a frequency parameter $\gamma_i$.  That is, $\gamma_i = \Pr_{\sample \drawnFrom \sampleDist}\left[\sample \in \group_i\right]$.  Parameter $\gamma$ is defined to be the minimum of all group frequencies: $\gamma = \min_{i} \gamma_i$.  We follow the lead of existing multicalibration literature and assume population groups may be overlapping \cite{shabat2020sample,hebert2017calibration}.  When referring to a general population group, we may use $\group$ without subscript.

% We now discuss our notation for population groups: given set of examples $\trainingData$, which will be clear from the context. 

A \emph{protected} feature is defined to be a feature to which our learner is blind. Though blind to the learner, such protected features may be of interest to society or an institution.  In this work, the population group $\group$ to which an example $\sample$ belongs is that example's protected feature.  That is to say, given an example $\sample$, the learner (i.e. trained classifier) is \emph{blind} to which population group $\group$ the example $\sample$ belongs, but the practitioner is \emph{aware} of the population group to which the example belongs.

Population groups are used within the definitions of category and calibration error.  More formally, 

\begin{definition}
    A category is a pair $(\group,\prediction)$ of population group $\group \in \groups$ and a predicted label $\prediction \in \labelSpace$.  

\end{definition}

% There are two parameters we use to characterize the behavior of categories. They are $\gamma$ and $\psi$.  Because $\gamma$ assume both  $(\group,\prediction)$, with respect to a predictor $\hyp$, is governed by two parameters $\gamma, \psi \in (0,1]$.   

The examples within training data which are members of category $(\group,\prediction)$ are denoted $\trainingData_{(\group,\prediction)}$.  That is

\begin{equation}
    \trainingData_{(\group,\prediction)} = \left\{\left(\sample,\labelElement\right) \in \trainingData \: :\: \sample \in \group, \labelElement=\prediction  \right\}
\end{equation}

The frequency with which a predictor $\hyp$ renders prediction $\prediction$ for each category $\group_i$ is denoted by $\psi_{\hyp,\prediction,i}$.  More precisely, $\psi_{\hyp,\prediction,i} = \Pr_{\sample \drawnFrom \sampleDist}\left[{\hyp\paren*{\sample} = \prediction \given \sample \in \group_i}\right]$. We denote by $\psi_{\hyp}$ the minimum such frequency over all such predictions: $\psi_{\hyp} = \min_{\prediction,\group} \psi_{\hyp,\prediction,i}$.  Thus, for a predictor $\hyp$, the frequency with which a category appears is $\gamma_i\psi_{\hyp,\prediction,i}$

Within the training regime, there are two losses with which we are concerned:  One is the training loss, which is typically a convex surrogate for accuracy such as cross-entropy loss or squared error.  The other is the calibration error which is the parity metric through which we understand fairness. In this work, we investigate how the empirical calibration error concentrates around the true multicalibration error.

  \begin{definition}[True Calibration Error]
    Let $\sampleDist \subseteq \sampleSpace \times \labelSpace$ be a sample distribution and let  $(\group,\prediction)$ be a category.  The calibration error of a predictor $\hyp \in \hypothesisClass$ with respect to a category $(\group,\prediction)$ is denoted $\calibrationError\paren*{\hyp,\group,\prediction}$:
    \begin{equation}
        \calibrationError\paren*{\hyp,\group,\prediction} \definedAs \E_{\paren*{\sample,\labelElement}\drawnFrom\sampleDist}\left[\hyp\paren*{\sample} \given \sample \in \group, \hyp\paren*{\sample}= \prediction\right] - \E_{\paren*{\sample,\labelElement}\drawnFrom\sampleDist}\left[{\labelElement \given \sample \in \group, \hyp\paren*{\sample}= \prediction}\right]
    \end{equation}
\end{definition}

The calibration error is the difference between the expectations of $\labelElement$ and $\hyp\paren*{\sample}$, conditioned on samples from population group $\group$ and the hypothesis $\hyp$ rendering a prediction $\prediction$ for example $\sample$.

\begin{definition}[Empirical Calibration Error]
\label{def:emp_cal_error}
    Let $\trainingData$ be a training set of $\numSamples$ examples drawn i.i.d. from sample distribution $\sampleDist\subseteq \sampleSpace \times \labelSpace$ and let $(\group,\prediction)$ be a category.  The empirical calibration error of a predictor $\hyp \in \hypothesisClass$ with respect to category $(\group,\prediction)$ and sample $\trainingData$ is denoted $\empiricalCalibrationError\paren*{\hyp,\group,\prediction,\trainingData}$

    \begin{equation}
        \empiricalCalibrationError\paren*{\hyp,\group,\prediction,\trainingData} \definedAs \sum_{(\sample,\prediction) \in \trainingData}\left\{\frac{\indicatorFunction\left[\sample \in \group, \hyp(\sample) = \prediction \right]}{\left|\trainingData_{(\group,\labelElement)}\right|}\times\left(\hyp(\sample)-\labelElement\right)\right\}
    \end{equation}
    
\end{definition}

Unlike the true calibration error, an ML practitioner may compute the empirical calibration error without explicit access to the underling distribution $\sampleDist$.  Intuitively, the empirical calibration error captures the difference between the average, taken over samples in $\trainingData$, of $\labelElement$ and $\hyp\paren*{\sample}$, conditioned on samples from population group $\group$ on which predictor $\hyp$ maps an example $\sample$ to prediction $\prediction$. Within this paper, we abuse terminology and often use calibration error and multicalibration error interchangably.

% Informally, a predictor with group fairness guarantees should exhibit little discrepancy between its average classification error on any group and the error incurred by any example within the same group.  This discrepancy is denoted as the calibration error. The true calibration error establishes this notion over the data distribution $\sampleDist$, while the empirical calibration error establishes a similar notion over the training set $\trainingData$. 

We focus on how empirical calibration error concentrates around the true calibration error for each population group. In particular, we focus on sample-complexity bounds. In the context of ERM, sample-complexity bounds on a hypothesis $\hyp \in \hypothesisClass$ provide the number of i.i.d. samples drawn from a distribution $\sampleDist$ necessary to estimate, with a sufficiently small error $\errorTolerance$ and suitably high probability $1-\errorConfidence$, the true risk $\risk_{\sampleDist}(\hyp)$ of predictor $\hyp$ over that distribution $\sampleDist$. It is defined as follows.

\begin{definition}[Sample Complexity Bound]
Given a loss function, error tolerance $\errorTolerance \in [0,1)$ and confidence parameter $\errorConfidence \in (0,1]$, a sample complexity $\sampleComplexity_{\hypothesisClass}(\errorTolerance,\errorConfidence)$ is the minimum number of samples such that the following holds:

\begin{equation}
    \Pr\left\{
    \left|\risk_{\sampleDist}(\hyp) \hiderel{-} \risk_{\trainingData}(\hyp) \right| \hiderel{\leq} \errorTolerance \right\} \geq 1-\errorConfidence
\end{equation}
\end{definition}

It is important to note that in some contexts, $\sampleComplexity_{\hypothesisClass}(\errorTolerance,\errorConfidence)$ is a \emph{distribution specific} bound. This means that it has a dependence on the data distribution, $\sampleDist$. 

For our purposes, it will also be useful to study the sample complexity for empirical risk convergence over a single population group, $\group \in \groups$. 

\begin{definition}[group sample complexity bound]\label{defn:group_complexity}
Let $\group \in \groups$ be a population group. Then given a loss function, error tolerance $\errorTolerance \in [0,1)$ and confidence parameter $\errorConfidence \in (0,1]$, a group sample complexity $\sampleComplexity_{\hypothesisClass, \group}(\errorTolerance,\errorConfidence)$ is the number of samples such that the following holds:

\begin{equation}
    \Pr\left\{
    \left|\risk_{\sampleDist_\group}(\hyp) \hiderel{-} \risk_{\trainingData_\group}(\hyp) \right| \hiderel{\leq} \errorTolerance \right\} \geq 1-\errorConfidence
\end{equation}

Here $\sampleDist_\group$ denotes the data distribution conditioned on membership in group $\group$, and $S_\group$ denotes a sample drawn from this conditional distribution.
\end{definition}

Sample-complexity bounds for multicalibration uniform convergence capture how the difference between the true and empirical calibration errors relate to a number of context dependent factors including the size of the training set, the number of categories, and the frequency of the least common population group. Such sample-complexity bounds yield high-probability guarantees that the true calibration error is similar to the empirical calibration error for all specified categories within a dataset. Our goal is to determine a bound on that sample complexity, denoted as $\sampleComplexity_{\hypothesisClass}^{\multicalibration}(\errorTolerance,\errorConfidence,\gamma,\psi)$.  That is, we wish to determine a bound on the number of samples needed to yield guarantees of the form:

\begin{equation}\label{eq:multicalibrationUniformConvergence}
    \Pr\left[\left\{\forall \hyp \hiderel{\in} \hypothesisClass,\forall \group \hiderel{\in} \groups, \forall \predictionValue \hiderel{\in} \labelSpace \hiderel{:} \\
    \left|\calibrationError(\hyp,U,\prediction) \hiderel{-} \hat{\calibrationError}(\hyp,U,\prediction,\trainingData) \right| \hiderel{\leq} \errorTolerance \right\} \right]\geq 1-\errorConfidence
\end{equation}

where $\errorTolerance,\errorConfidence,\psi,\gamma \in (0,1]$.

Our method for doing so will be to bound the multicalibration complexity using the group sample complexity bound defined in Definition \ref{defn:group_complexity}.

\section{Relating Calibration and ERM Generalization}
\label{sec:our_bound}

With notation defined, we now focus on our contribution regarding multicalibration uniform convergence sample complexity bounds. The main contribution of this section is the ease with which such sample complexity bounds may be devised: \textit{Multicalibration uniform convergence sample complexity bounds are a reparametrization of ERM learning sample complexity bounds.} We begin with a theorem which formalizes the relationship between ERM and Multicalibration.  The theorem has implications on the relationship between multicalibration, hypothesis class complexity, and dataset balancing.

\begin{theorem}\label{thm:maintheorem}
   Let $\trainingData$ be an i.i.d. sample drawn from distribution $\sampleDist$, let $\groups$ be the set of groups, and let $\gamma,\psi$ be the frequency parameters described in \cref{sec:background}.  Then the multicalibration complexity can be bounded with the maximum group sample complexity. That is,  
   
  \begin{equation}
      \sampleComplexity_{\hypothesisClass}^{\multicalibration}(\errorTolerance,\errorConfidence,\gamma,\psi) \leq \max_{\group \in \groups} \frac{2}{\gamma}\sampleComplexity_{\hypothesisClass, \group}\left(\frac{\psi\errorTolerance}{3},\frac{\errorConfidence}{4\abs{\groups}\abs{\labelSpace}}\right).
  \end{equation}
\end{theorem}

\textbf{Proof Ideas:} The main idea of proving Theorem \ref{thm:maintheorem} is to decompose the multicalibration problem into $\frac{1}{|\groups||\labelSpace|}$ separate instances. Each instances corresponds to a category $(\group, y)$. Then, for each category, we show that multicalibration for the category can be achieved by using standard ERM bounds \textit{for that category.} In particular, the category $(\group, \prediction)$ can be bounded in terms of $\sampleComplexity_{\hypothesisClass, \group}$. We then determine the number of samples needed so that each demographic group will have a sufficient number of samples to draw from (providing the $\frac{2}{\gamma}$ term). A full proof of this theorem appears in \cref{sec:ProveMainTheorem}.

% {\color{red} We may wish to make the above definition more abstract? The bounds may depend on training regime more broadly than just the hypothesis class and sample distribution, but also a myriad of factors we may wish to address. }

This theorem implies that group sample complexity is \textit{sufficient} for multicalibration. Note that these are two distinct tasks: group sample complexity is the amount of data needed to correctly estimate the loss of a given classifier whereas multicalibration is the amount of data needed to correctly estimate biases conditioned on \textit{category.} Furthermore, this theorem implies that any bounds that can be instantiated for empirical risk minimization will imply bounds on multicalibration.

Demographic (im)balance within the dataset has been well discussed in fairness literature \cite{dwork2012fairness,buolamwini18a,barocas-hardt-narayanan}.  Interestingly, \cref{thm:maintheorem} bounds have implications in dataset balancing as well. Demographic balance manifests itself as $\gamma$. For fixed $\hypothesisClass,\errorTolerance,\errorConfidence,\psi$ and $\groups$, sample complexity for multicalibration uniform convergence $\sampleComplexity_{\hypothesisClass}^{\multicalibration}(\errorTolerance,\errorConfidence,\gamma,\psi)$ is minimized when $\gamma$ is maximized. This occurs when $\gamma = \frac{1}{\abs{\groups}}$, which is when the number of examples in each demographic is the same, i.e. when the dataset is demographically balanced.

The predictor class complexity factors into the multicalibration uniform convergence sample complexity bounds via the ERM group sample complexity bound $\sampleComplexity_{\hypothesisClass, \group}(\errorTolerance, \errorConfidence)$. The group sample complexity bound itself is a standard ERM sample complexity bound $\sampleComplexity_{\hypothesisClass}(\errorTolerance, \errorConfidence)$ conditioned on all samples being members of the specified group $\group$. In general it takes fewer samples to learn a less complicated predictor class than it does to learn a more complicated predictor class. Consider, for example, the sample complexity of a predictor class $\hypothesisClass$ with Vapnik-Chervonenkis (VC) dimension $\graphDimension$. Shalev-Shwartz and Ben-David showed that $\sampleComplexity_{\hypothesisClass}(\errorTolerance, \errorConfidence) = \bigTheta\left(\frac{\graphDimension+\log\left(\sfrac{1}{\errorConfidence}\right)}{\errorTolerance^2}\right)$ \cite{shalev2014understanding}.  For fixed $\errorTolerance,\errorConfidence,\gamma,\psi$ and number of groups $\abs{\groups}$, as the VC dimension \cite{vapnik1971uniform} of a hypothesis class increases, so does the multicalibration uniform convergence sample complexity $\sampleComplexity_{\hypothesisClass}^{\multicalibration}(\errorTolerance,\errorConfidence,\gamma,\psi)$. For the sake of completeness, VC dimension is defined in \cref{sec:boundsSec3}.

\subsection{Illustrative Examples}

The choice of ERM sample complexity term also allows machine learning practitioners to loosen or tighten multicalibration uniform convergence bounds depending on how much information about the underlying classification scenario.  The more information a practitioner has, the sharper their bounds can be. For illustrative purposes, let us consider the following three scenarios. 

\begin{description}
	 \item[Scenario 1] The practitioner is aware of the predictor class $\hypothesisClass$, but has yet to receive any data.  
    \item[Scenario 2] The practitioner is aware of both the predictor class $\hypothesisClass$ and has received their labeled training dataset $\dataset$.
    \item[Scenario 3] The practitioner is aware of the predictor class, training sample $S$, and has additional knowledge of the data distribution.
\end{description}

These situations are ordered in increasing specificity -- each subsequent situation gives the practitioner more information which allows them to make tighter estimates on the amount of data needed for multicalibration. 

\subsubsection{Scenario 1: VC Dimension}

In this first setting, the practitioner has enough information to invoke the VC-dimension based ERM learning sample-complexity bounds. VC-dimension is the canonical distribution agnostic tool used to bound the sample complexity for learning a hypothesis class $\hypothesisClass.$ In particular, if $\hypothesisClass$ has VC-dimension $p$, then 
\begin{equation}
\sampleComplexity_{\hypothesisClass}(\errorTolerance, \errorConfidence) = \bigOh \left(\frac{p + \log\frac{1}{\delta}}{\epsilon^2} \right)
\end{equation}

As such, the practitioner can use \cref{thm:maintheorem} to recover the bounds of Shabat et al \cite{shabat2020sample}. That sample complexity bound is

\begin{equation}
    \sampleComplexity_{\hypothesisClass}^{\multicalibration}(\errorTolerance,\errorConfidence,\gamma,\psi) = \bigOh \left(\frac{1}{\errorTolerance^2\psi^2\gamma}\left(\graphDimension+\log\left(\frac{\abs{\groups}\abs{\labelSpace}}{\errorConfidence}\right)\right)\right)
\end{equation}

\subsubsection{Scenario 2: Rademacher Complexity}

Because the practitioner now has access to both dataset $\dataset$ and the hypothesis class itself, a Rademacher-complexity based ERM learning sample-complexity bound may be invoked.  Some predictor classes have finite Rademacher complexity, but infinite or otherwise undefined VC-dimension.  The most notable such hypothesis class is the Radial Basis Function (RBF) kernel SVM. Hence, the ability to invoke a Rademacher complexity ERM learning bound may grant the ability to construct a multicalibration uniform convergence bound in a scenario for which a VC-dimension is uninformative. For the sake of completeness, Rademacher complexity is defined in \cref{sec:rad_appendix}.

We now illustrate this more precisely with two examples: Kernel SVM predictors, and ReLU activated neural networks.

\paragraph{RBF Kernels:} Consider the Kernel SVM predictor class $\hypothesisClass_{\kernel}$:

\begin{equation}\label{eq:kernelHypClass}
 \hypothesisClass_{\kernel} \definedAs \left\{\sample \mapsto \sgn\left(\weights^{\transpose}{}\kernelMapping_{\kernelFunction}(\sample)\right) : \abs{\weights^{\transpose}{}\kernelMapping_{\kernelFunction}(\sample)} \geq 1\right\}   
\end{equation}

 where  $\kernelMapping : \realNumbers^\dimension \to \realNumbers^{\dimension'}$ is the kernel mapping function and $\dimension' \in \naturalNumbers \union \infty$.  Often, the dimension of the kernel $\dimension'$ is exceedingly large, so the \emph{kernel trick} is utilized to reduce the computational complexity of both the training of and prediction with a kernel predictor class. The kernel trick takes predictor classes of the form depicted in \cref{eq:kernelHypClass} and represents them as follows:
 
\begin{equation}\label{eq:kernelPredictor}
    \hyp(\sample) = \sum_{i=1}^\numSamples \weight_i\labelElement_i \kernelFunction\left(\sample_i,\sample \right)
\end{equation}
 
  Notably, for the RBF Kernel, $\dimension'$ is $\infty$.  Hence, the kernel trick is the only way to pragmatically deploy an RBF kernel.

By computing the Rademacher complexity (details are in section \ref{sec:boundsSec3} for this hypothesis class), we have that
\begin{equation}
\sampleComplexity_{\hypothesisClass_{\kernel}}(\errorTolerance, \errorConfidence) = \bigOh \left(\frac{B^2 + \log \frac{1}{\delta}}{\epsilon^2} \right),
\end{equation}

where $\scalarBound$ denotes the maximum value of $\kernel(\sample,\sample)$ over all $\sample$. This bound holds uniformly over all groups as the value of $\scalarBound$ is unchanged between groups. Substituting this into Theorem \ref{thm:maintheorem}, we find a multicalibration complexity of 

\begin{equation}
\sampleComplexity_{\hypothesisClass_{\kernel}}^{\multicalibration}(\errorTolerance,\errorConfidence,\gamma,\psi) = \bigOh\left(\frac{1+ \log\left(\frac{\left|\Gamma\right|\left|\labelSpace\right|}{\errorConfidence}\right)}{\errorTolerance^2\psi^2\gamma}\right)
\end{equation}

\paragraph{ReLU Networks:} We can also devise a sample complexity bound for ReLU networks. Before providing the bound, we provide notational setup.  The predictor class, denoted by $\hypothesisClass_{\neuralNetwork}$, is parametrized by a sequence of matrices $\weightMatrix = (\weightMatrix_1,\dotsc,\weightMatrix_\depth)$, where $\weightMatrix_{i} \in \realNumbers^{\dimension_{i}\times\dimension_{i-1}}$ and $\dimension_0 = \dimension$. The depth of the network is $\depth$. In the binary classification setting, we have $\dimension_{\depth} = 1$, and in the multi-class setting, we have $\dimension_{\depth} = \abs{\labelSpace}$.  Let $\nonlinearity(\cdot)$ be the ReLU function.  $\nonlinearity(\vec{\placeholder})$ applies the ReLU to each coordinate in $\vec{\placeholder}$. Such networks are of the form
\begin{equation*}
    \weightMatrix_{\depth}\nonlinearity\left(\weightMatrix_{\depth-1}\nonlinearity\left(\cdots\nonlinearity\left(\weightMatrix_{1}\sample\right)\cdots\right)\right)
\end{equation*}

Thus, the ReLU predictor class is
\begin{equation}\label{eq:neuralNetworkHypothesisClass}
    \hypothesisClass_{\neuralNetwork} = \left\{\hyp_{\weightMatrix}(\sample) : \weightMatrix : \left(\weightMatrix_1,\dotsc,\weightMatrix_\depth\right), \norm{\weightMatrix_i}_2 \leq \placeholder_i, \norm{\weightMatrix_i^{\transpose}}_{2,1} \leq \placeholder'_{i}, i \in \left\{1,\dotsc,\depth\right\}\right\}
\end{equation}

By bounding the Rademacher complexity of this hypothesis class, under some mild assumptions we find a standard sample complexity (details in Appendix \ref{sec:boundsSec3}) satisfying
\begin{equation}
    \sampleComplexity_{\hypothesisClass_{\neuralNetwork}}(\errorTolerance, \errorConfidence) \leq \frac{7200\log^2\left(2\dimension_{\max}\right)\norm{\vec{X}}_{\Frobenius}^2\left(\prod_{i=1}^{\depth}\placeholder_i\right)^2\left(\sum_{j=1}^{\depth}\left(\frac{\placeholder_j'}{\placeholder_j}\right)^{\sfrac{2}{3}}\right)^{3} + 64\log\left(\sfrac{4}{\errorConfidence}\right)}{\errorTolerance^2}
\end{equation}

This bound holds uniformly over all groups $g$ as the key parameter $\norm{\vec{X}}_{\Frobenius}^2$ upper bounds the corresponding quantity for all groups. Substituting this, the multicalibration uniform convergence sample complexity bound for this predictor class is
\begin{equation}
\sampleComplexity_{\hypothesisClass_{\neuralNetwork}}^{\multicalibration}(\errorTolerance,\errorConfidence,\gamma,\psi) = \bigOh\left(\frac{\log^2\left(\dimension_{\max}\right)\norm{\vec{X}}_{\Frobenius}^2\left(\prod_{i=1}^{\depth}\placeholder_i\right)^2\left(\sum_{j=1}^{\depth}\left(\frac{\placeholder_j'}{\placeholder_j}\right)^{\sfrac{2}{3}}\right)^{3} + \log\left(\frac{\left|\groups\right|\left|\labelSpace\right|}{\errorConfidence}\right)}{\errorTolerance^2\psi^2\gamma}\right) 
\end{equation}

where $\vec{X}$ is the unlabeled portion of a dataset $\trainingData$, i.e. $\vec{X} = \left[\sample_1,\dotsc,\sample_N\right]$. 

\subsubsection{Scenario 3: Distribution specific bounds}

In this case, the practitioner is able to utilize their full knowledge of the distribution to obtain even tighter bounds. One important example of this is hard margin SVM classification. In this setting, let $\hypothesisClass$ be the set of all linear classifier, and let $\sampleDist$ be any distribution with hard margin $\rho$. The existence of a hard margin guarantees the existence of a linear classifier which achieves perfect accuracy on $\sampleDist$ with a margin of $\rho$, meaning that any point within distance $\gamma$ of the given point will also be classified the same. 

In this case, it is well known that $\sampleComplexity_{\hypothesisClass}(\epsilon, \delta) = O\left(\frac{D^2\log(\frac{1}{\delta})}{\rho^2 \epsilon}\right)$ where $D$ denotes the distance between the furthest two points in the data distribution. 

In our case, this bound will still hold for all groups, as the margin and diameter for each group must be at least $\rho$ and at most $D$ for each $\group \in \groups$. Applying the main theorem, we have that 

\begin{equation}
    \sampleComplexity_{\hypothesisClass}^{\multicalibration}(\errorTolerance,\errorConfidence,\gamma,\psi) = \bigOh \left( \frac{D^2\log(\frac{4|\groups||\labelSpace|}{\delta})}{\rho^2 \gamma \psi \epsilon}\right).
\end{equation}

Observe that this bound is significantly tighter than the bounds in previous sections -- there is no dependence on any sample complexity measures outside of $\rho$ and $D$. Notably, $\rho$ and $D$ are both distances and have no dependence on the dimension of the data distribution. By contrast, both VC dimension and Rademacher complexity tend to include a term related to the dimensionality of the data.

\section{Evaluation}
\label{sec:evaluation}

Our evaluation aims to show the convergence behavior of the multicalibration error. Towards that end, we leverage three popular datasets in ML experiments: the UCI Adult dataset, the Correctional Offender Management Profiling for Alternative Sanctions (COMPAS) from ProPublica, and the CelebA face recognition dataset from the Chinese University of Hong Kong. Three questions guide our evaluation:
\begin{description}

\item [{Q1.}] How does the data distribution affect the baseline multicalibration error dispersion?
    
\item [{Q2.}] Does the data distribution affect the convergence behavior of multicalibration error?

\item [{Q3.}] Does the choice of hypothesis class impact the convergence behavior of multicalibration error?
\end{description}

\subsection{Experimental Setup}
\label{sec:experiment_setup}

The purpose of experiments is to study the impact of data distribution and hypothesis class on multicalibration uniform convergence behavior. In an ideal world, we would explicitly plot, for each classifier, the convergence of empirical multicalibration error to true multicalibration error, i.e., $\calibrationError(\hyp,\group,\prediction)-\empiricalCalibrationError(\hyp,\group,\prediction,\trainingData)$, as a function of category frequency $\gamma_i\psi_{\hyp,\prediction,i}$ and/or the number of examples in the dataset. Unfortunately, this requires access to the underlying data distribution, which we do not have.

Since we do not have access to the underlying data distribution, we simulate draws from it. This procedure is inspired by the seminal work of Zhang et al. \cite{zhang2016understanding}, in which randomness is applied to break the connection between examples and their labels. Instead of decoupling examples from their labels, we use randomness to decouple classifier performance from the explicit composition of training datasets. To do so, we generate many unique train-test splits, and we train classifiers on each such dataset split. 

Let $\left[\dataset_{\train}, \dataset_{\test}\right]$ by a partition of dataset $\trainingData$. Given the dataset $\dataset$, groups $\group_1$ and $\group_2$ (not necessarily the only groups in the dataset), and list of predictor classes, let $\placeholderSet_{\group_1},\placeholderSet_{\group_2}$ be sets denoting the number of examples in $\dataset_{\train}$ belonging to groups $\group_1$ and $\group_2$, respectively. Further, let $\placeholder_1,\placeholder_2$ denote elements of  $\placeholderSet_{\group_1}$ and $\placeholderSet_{\group_2}$, respectively. For each unique ordered pair $\placeholder_1,\placeholder_2$, we sample uniformly at random without replacement $\placeholder_1$ examples from $\group_1$ and $\placeholder_2$ examples from $\group_2$. $\dataset_{\train}$ consists of these randomly sampled examples along with all examples in $\dataset$ which are neither in group $\group_1$ nor in group $\group_2$. $\dataset_{\test}$ is the complement of $\dataset_{\train}$.  For each such train-test split, we train one classifier of each predictor class on training data $\dataset_{\train}$ and that classifier is evaluated on $\dataset_{\test}$.  We repeat this procedure 25 times for each such unique ordered pair $\placeholder_1,\placeholder_2$.

This procedure allows us to construct balanced datasets, and datasets that are imbalanced but with known demographic proportions. Because the test dataset $\dataset_{\test}$ is the complement of $\dataset_{\train}$, controlling the demographics of our training set means also means we are controlling that of our test set. Consequently, this procedure allows us to simulate, with finite data, the convergence behavior of empirical calibration error of each category.

\subsection{Datasets and Classifiers}

We discuss the evaluation datasets and their preparation. Categorical data are one-hot encoded, while numerical and ordinal data are left unmodified. The protected attributes for UCI Adult and COMPAS are race and sex, and the protected attribute for CelebA is sex. We plot, for each category, the empirical multicalibration error ($y$-axis) against the empirical category frequency ($x$-axis). For each train-test split and its associated classifier, the calibration error of each category is represented as a single dot on the associated scatterplot.

\subsubsection{Adult Dataset.}
\label{subsubsec:Adult_setup}
        
The Adult dataset, from the UCI machine learning repository, comprises 14 attributes and 48842 records. Entries in the dataset are extracted from the 1994 Census database. The label indicates if the income of a person exceeds \$50K. We assign an individual earning over \$50K the label $\labelElement = 1$, and if they do not, we assign them a label $\labelElement=0$.  Classifiers for the Adult dataset are trained on the following features: \texttt{Age}, \texttt{Workclass}, \texttt{fnlwgt}, \texttt{Education-Num}, \texttt{Martial Status},
        \texttt{Occupation}, \texttt{Relationship}, \texttt{Capital Gain}, \texttt{Capital Loss},
        \texttt{Hours per week}, and \texttt{Country}.  The sex and race features are withheld from training and are reserved as protected attributes.  
        
        When implementing the procedure previously defined, the two population groups, $\group_1$ and $\group_2$, correspond to male and female, respectively. Results pertaining to the race attribute may be found in \cref{sec:AdditionalFigures}. It is worth noting that the dataset contains 16192 entries representing females and 32650 entries representing males. We use $\placeholderSet_{\mathrm{male}} = \placeholderSet_{\mathrm{female}} = \left\{200,400,800,1600,3200,6400,12800\right\}$. When plotting, we only consider the classifiers trained using training sets of size between 4500 and 9000 examples. For each train-test split, we train a RBF Kernel SVM, Linear SVM, and a two-layer ReLU network consisting of 1000 hidden units.

\subsubsection{COMPAS Dataset.}
\label{subsubsec:Compas_setup}
        
The COMPAS dataset from ProPublica contains the features utilized by the COMPAS algorithm in assigning the risk of reoffending for defendants in Broward County, Florida. The dataset comprises roughly 7200 records and 28 features; the label indicates if a defendant re-offends within two years. If the defendant re-offends, we assign a label of $\labelElement = 1$, and if they do not, we assign a label of $\labelElement=0$. Classifiers on the COMPAS dataset are trained upon the following six features:  \texttt{age}, \texttt{juv_fel_count}, \texttt{juv_misd_count}, \texttt{priors_count}, \texttt{charge_id}, and \texttt{charge_degree (misd/fel)}. Sex and race features are withheld from training and are reserved as protected attributes.

The two population groups $\group_1$ and $\group_2$ correspond to male and female, respectively. Results pertaining to the race attribute may be found in \cref{sec:AdditionalFigures}. It worth noting that the dataset contains 1395 females and 5819 males. We use $\placeholderSet_{\mathrm{male}} = \placeholderSet_{\mathrm{female}} = \left\{20,40,80,160,320,640,1280\right\}$. 
When plotting, we only consider the classifiers trained using training sets of size between 800 and 1600 examples. For each train-test split,  we train an RBF Kernel SVM, Linear SVM, and a two-layer ReLU network consisting of 1000 hidden units.

\subsubsection{CelebA Dataset.}
\label{subsubsec:CelebA_setup}

The CelebA \cite{liu2015faceattributes} dataset consists of 202599 RGB face images from a total of 10177 identities.  There are 40 labeled binary attributes for each image, including sex, presence of mustache, presence of straight hair, and whether or not someone is smiling. If an attribute is present or otherwise holds true, we assign it a label $\labelElement=1$, and if not,  we assign it a label $\labelElement=0$. We use the sex attribute as a protected feature. We do not train classifiers on the CelebA data using raw images. Instead, we use 2048 dimensional embeddings of images obtained from a ResNet-152 \cite{he2016resnet} pretrained on the ImageNet \cite{Deng09imagenet} dataset.

Since we only have access to the sex attribute for examples in CelebA, we only construct train-test splits when conditioning on the sex attribute.  Plots are generated for 
classifiers trained using training sets of size between 27000 and 54000 images. When generating train-test splits, we have $\placeholderSet_{\mathrm{male}} = \placeholderSet_{\mathrm{female}} = \left\{200,400,800,1600,3200,6400,12800,25600,51200\right\}$. For each train-test split,  we train both a Linear SVM and a two-layer ReLU network consisting of 1000 hidden units.

Note that face recognition datasets, including CelebA, have become important benchmarks for machine learning practitioners. Such face recognition datasets have recently come under scrutiny due to ethics and privacy concerns. Many of these concerns relate to discrimination (racism, sexism, etc.) \cite{barocas2016big}. Consequently, dataset management procedures have been put forth to help the community become responsible dataset stewards \cite{hutchinson2021towards}. Given that CelebA is a widely disseminated dataset with strong precedent in existing literature, we feel our use of CelebA has benefit that outweighs any ethical or privacy concern. Furthermore, we do not make broad-scale conclusions regarding demographic groups: we only measure and characterize the convergence behavior of calibration error.

While there are multiple possible values for the race attribute, the Black and White groups are by far the most dominant groups in both UCI Adult and COMPAS. Hence, when conditioning on the race attribute, we augment the training set by including examples that are neither black nor white. Since there are only two sexes in the three datasets, such augmentation is unnecessary when conditioning by sex.

 \subsection{Evaluation Results}

\begin{figure}[t]
    \centering
        \includegraphics[width=1.0\textwidth]{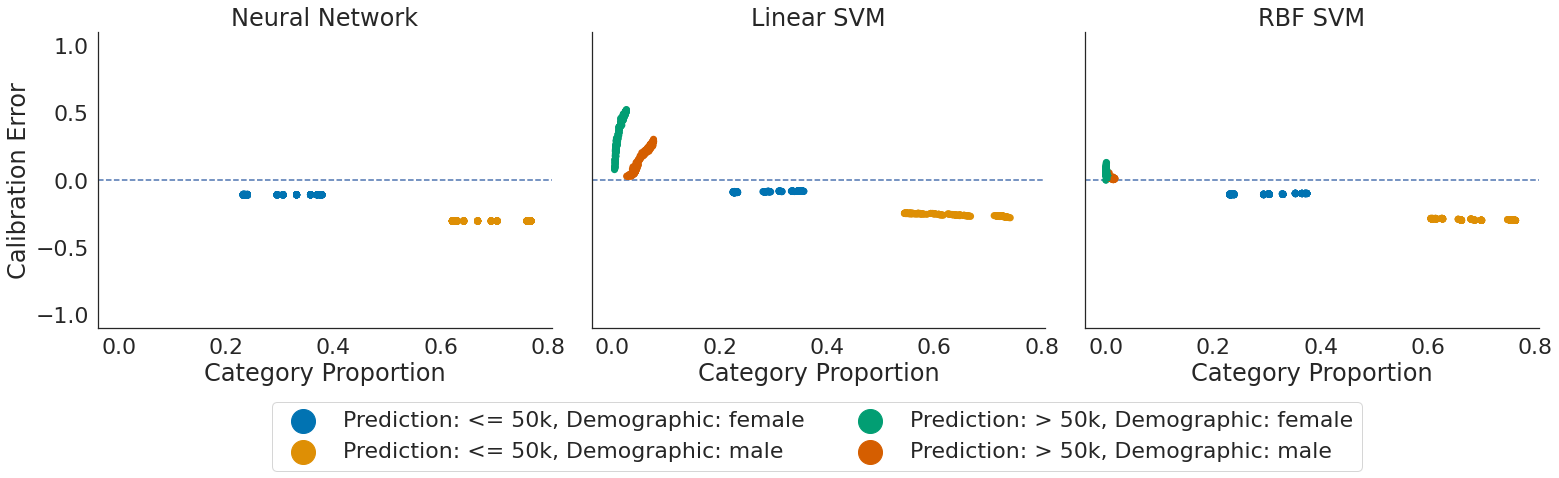}
    \caption{The multicalibration convergence behavior on the UCI Adult dataset, conditioned by sex.}
    \label{fig:AdultSex}
\end{figure}

\begin{figure}[t]
    \centering
    \includegraphics[width=1.0\textwidth]{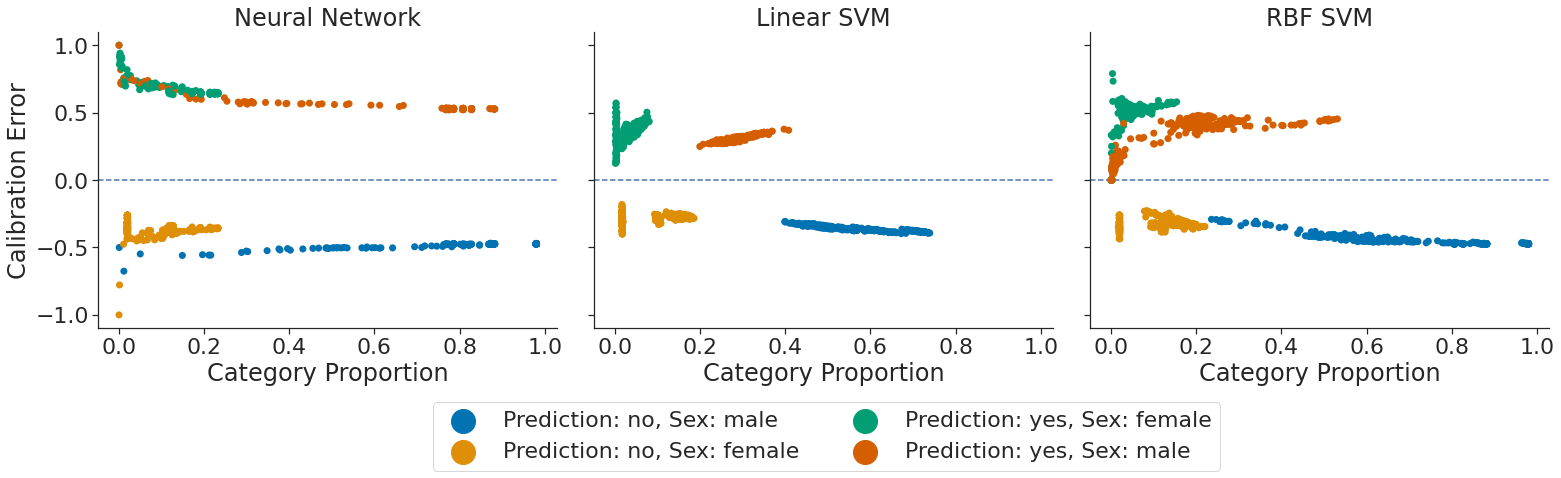}
    \caption{The multicalibration convergence behavior on the COMPAS dataset, conditioned by sex. }
    \label{fig:CompasSex}
\end{figure}

\begin{figure}[t]
    \centering
    \includegraphics[width=1\textwidth]{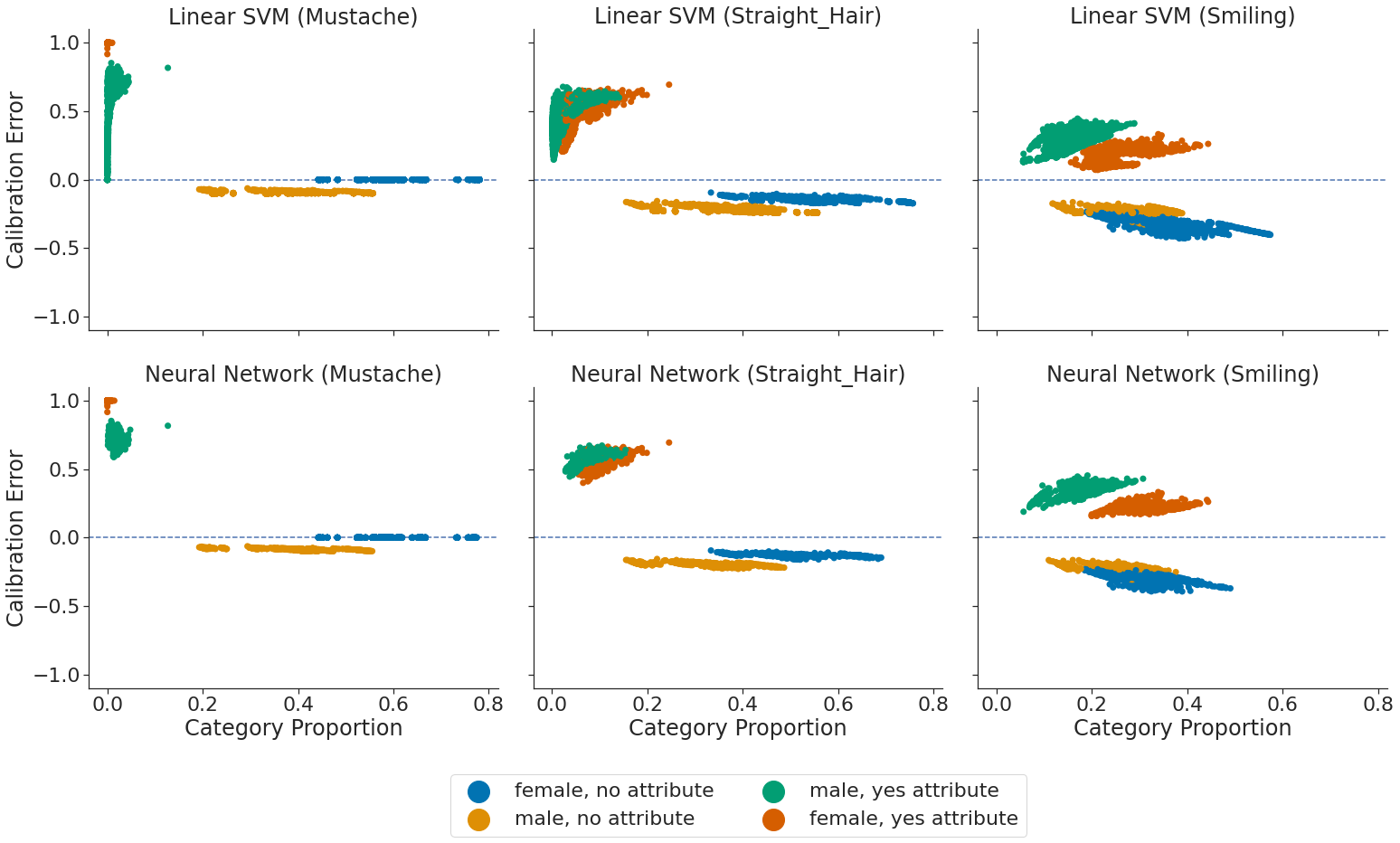}
    \caption{The convergence behavior of calibration error for the CelebA dataset, conditioned by sex. Note that the clusters for each category can overlap, obscuring the true shape of the background cluster. 
    For both the linear SVM and the neural network, the cluster in which males are predicted to have straight hair overlaps the cluster representing females also predicted to have straight hair. Also, for both the linear SVM and the neural network, the cluster representing females predicted not to be smiling obscures the cluster representing males predicted not to be smiling. }
    \label{fig:celeba}
\end{figure}

Each of \cref{fig:AdultSex,fig:CompasSex,fig:celeba} captures the multicalibration error of multiple predictor according to the procedure defined in \cref{sec:experiment_setup}. These plots represent scatterplots, where each point corresponds to the calibration error of a given classifier, category pair. The $x$-coordinate is the category's frequency $\gamma_i\psi_{\hyp,\prediction,i}$ in the given classifier's test set $\dataset_{\test}$, and the $y$-coordinate is that category's empirical calibration error. We can grasp the empirical distribution of multicalibration error by displaying all such points.

The plots in \cref{fig:AdultSex,fig:CompasSex,fig:celeba} allow us to understand how the multicalibration error convergence rate relates to each dataset, its constituent demographics, and the predictor classes upon which learning occurs. This convergence rate is interpreted by visual inspection: A faster convergence rate means the vertical dispersion of calibration error, a proxy for $\errorTolerance$ (the difference between true calibration error $\calibrationError$ and empirical calibration error $\empiricalCalibrationError$), shrinks more quickly as category proportion increases. Existing ERM bounds provide the best lens of hypothesis class complexity through which we may examine the relationship between multicalibration error convergence and \cref{thm:maintheorem}.\\

\noindent
\textbf{Q1.} \textit{How does the data distribution affect the baseline multicalibration error dispersion?}

Understanding the baseline level of dispersion in multicalibration error is an important foundation for discussing its convergence behavior. This dispersion is directly related to the number of samples in each dataset. In the smallest dataset, COMPAS, we see the largest dispersion in calibration error. This is depicted in \cref{fig:CompasSex}.  At the other end of the spectrum is CelebA, our largest dataset, which is depicted in \cref{fig:celeba} has the least dispersion. When considering the dispersion properties of the Adult dataset, it is important to note the vast imbalances in category proportions. Such imbalances lead to vertical-appearing trends suggesting a large dispersion, but indeed the solution to a linear regression drawn through each category would have non-infinite slope. Consequently, the meaning the vertical dispersion is much less than a naive glance might suggest. We conclude that for a fixed category proportion, the larger the dataset, the less is the dispersion in calibration error. Indeed, if we fix all parameters in our main theorem with the exception of $\errorTolerance$, we see that decreases in $\errorTolerance$ necessitate an increased sample complexity $\sampleComplexity_{\hypothesisClass}^{\multicalibration}(\errorTolerance,\errorConfidence,\gamma,\psi)$. \\

\noindent
\textbf{Q2.} \textit{Does the data distribution affect the convergence behavior of multicalibration error?}

 The behavior we see in \cref{fig:AdultSex,fig:CompasSex,fig:celeba} captures that of ERM bounds upon which our main theorem is based.  With respect to the frequency in which a category appears in a dataset, our experiments confirm the convergence behavior of calibration error suggested by our theory (\cref{thm:maintheorem}). Furthermore, as the frequency of a category increases in a dataset, we observe the multicalibration error tends to converge. In a significant departure from the behavior suggested by ERM convergence, there is a mean shift when plotting the category-wise calibration error. Despite this mean shift, convergence behavior of multicalibration error is present as our bounds indicated would be the case. With respect to error convergence behavior, we observe different convergence rates for each dataset. This behavior is related to the number of samples in each dataset. COMPAS, our smallest dataset, as depicted in \cref{fig:CompasSex}, has some of the strongest convergence behavior among its categories. For this dataset, calibration error for each category appears as a thin line. Whereas, for the CelebA dataset as plotted in \cref{fig:celeba}, dispersion in calibration error is apparent. The dispersion of multicalibration error for the Adult dataset lies between the two. We attribute this to our sampling strategy, with less randomness in drawing training sets from a smaller dataset. However, the calibration error in COMPAS converges (faster) away from zero as evident in \cref{fig:CompasSex}, indicating a structural unfairness issue, attributed to the small dataset size. \\

\noindent
\textbf{Q3.} \textit{Does the choice of hypothesis class impact the convergence behavior of multicalibration error?}

The choice of predictor class appears to have an impact on the convergence behavior of multicalibration error, but not in the manner our bounds suggest. For all three datasets, neural networks appear to have the sharpest convergence behavior in multicalibration error. While the behavior of multicalibration error in our bounds suggests significant dispersion, this does not translate in our experiments. We attribute this discrepancy to the overparametrization of neural networks combined with the notoriously loose nature of neural network information capacity bounds. Consequently, sample complexity bounds for multicalibration error convergence in neural networks are pessimistic. The behavior of multicalibration error on SVM predictor classes is sharper than that of the neural network. Linear SVMs are less expressive than RBF SVMs, and this behavior is best captured in the plots for COMPAS \cref{fig:CompasSex}. The convergence behavior of the Linear SVM is stronger than that of the RBF SVM. In the Adult dataset, the number of examples in the least frequent categories is insufficient to draw any substantive conclusions.

   % plot $\empiricalCalibrationError(\hyp,\group,\prediction,\trainingData_{\test})$ where $\trainingData_{\test}$ is the 

% Given a dataset $\trainingData$, we divide i

    % \begin{enumerate}
    %     \item With respect to population demographics, how does multicalibration error dispersion behave across hypothesis classes?  Is convergence slower or otherwise less evident when hypothesis classes are more complicated? 
    %     \item Are there disparities in multicalibration error dispersion with respect to population category frequency, as indicated by our bounds?
    %     % \item With respect to the total number of examples in the training set, are there disparities in multicalibration error dispersion as indicated by our bounds?  Does the behavior match that which we calculated in our bounds? This question is evaluated on COMPAS, CelebA, and UCI Adult datasets.
    % \end{enumerate}
\section{Related Work}
The following discusses the related works to sample complexity measures in ERM, multicalibration error, and other fairness notions. Further, we highlight our contributions relative to these related works.

\subsection{ERM Learning}

Our results demonstrate how sample complexity bounds for ERM learning translate into multicalibration convergence bounds. ERM bounds are common in literature, including bounds on decision trees, neural networks, and other families of classifiers. Given the prevalence of machine learning classifiers and associated bounds, our main theorem (\cref{thm:maintheorem}) makes the rendering of multicalibration sample complexity bounds accessible for any ML practitioner who uses ERM bounds.

Much of modern ML fairness literature is focused on an application of constrained ERM known as learning fair classifiers. This is accomplished by performing ERM, or some variant thereof, subject to what is known as a fairness constraint. A fairness constraint is often formulated as an optimization constraint and is dependent on population group $\group$, or another protected attribute.   Examples of fairness constraints include \emph{statistical parity} (also referred to as \emph{demographic parity}) introduced by Dwork et al. \cite{dwork2012fairness} and \emph{equality of opportunity} introduced by Hardt et al. \cite{hardt2016equality}.  These works differ from ours in that a learner is assumed to have access to protected attributes: they assume the fairness constraint(s) can be enforced during training.

\subsection{Multicalibration Error}

Multicalibration error is a well-studied quantity measuring group-wise performance parity. This notion is useful when classifiers do not have access to group information; Hebert-Johnson et al. were the first to develop and study the notion of multicalibration \cite{hebert2017calibration} in the context of machine learning. They develop several algorithms for learning multicalibrated predictors and analyze their sample complexity. 

Liu et al. consider a two-protected group binary decision problem. The learner aims to identify the selection policy, which maximizes utility subject to a fairness constraint. Samples are assumed to have a score that serves as a proxy for selecting the utility. The authors relate the multicalibration of a learned selection policy subject to a fairness constraint to the prediction utility \cite{liu2018delayed}.

Liu et al. also study the relationship between multicalibration and unconstrained machine learning \cite{liu2019implicit}. They show that for a large class of loss functions, if a learned classifier has a small excess risk over the Bayes risk when conditioning on the protected attribute(s), the learned classifier will also be well-calibrated.

More recently, Shabat et al. utilize the graph dimension \cite{daniely2011multiclass}, a multi-class generalization of the VC dimension, to determine the sample-complexity necessary for uniform convergence guarantees on multicalibration error \cite{shabat2020sample}. Their results have no dependence on the underlying learning algorithm. Jung et al. \cite{pmlr-v134-jung21a} design algorithms which yield predictors that have guarantees on higher central moments of multicalibration error. This provides information on the geometry of the multicalibration error beyond what is given by standard mean estimates.

In this work, we showed that multicalibration error sample complexity bounds are reparametrizations of existing ERM learning sample complexity bounds. We also put forth new Rademacher-complexity based multicalibration uniform convergence sample complexity bounds.

\subsection{Other Fairness Notions}

Related to the multicalibration error is the notion of Outcome Indistinguishability. Outcome Indistinguishability, as introduced by Dwork et al. \cite{dwork2021outcome}, measures the degree to which a set of examples as labeled by nature differs from the same set of examples labeled as a synthetic or trained classifier. Furthermore, Dwork et al. demonstrate that enforcing outcome indistinguishability transitively enforces multicalibration and vice-versa. Kearns et al. utilized the VC dimension \cite{vapnik1971uniform}  in their study of auditing algorithms for the prevention of fairness gerrymandering \cite{kearns2018preventing}. Another closely related paper is Rothblum and Yona's \cite{rothblum2021multi} study of Multi-group Probably Approximately Correct (PAC) Learnability. For a fixed loss function, the authors provide an algorithm that produces a classifier that guarantees that each population group has a similar average loss. The authors also provide sample complexity bounds for Multi-group PAC learning.

%!TeX root = ../note.tex

\section{Conclusion and Future Work}

This paper explored machine learning fairness from the perspective of sample complexities for multicalibration uniform convergence. We proved that multicalibration uniform convergence bounds are reparametrizations of ERM learning sample complexity bounds. This reparametrization subsumes existing state-of-the-art sample complexity bounds for multicalibration error uniform convergence. Furthermore, utilizing our main theorem, we put forth a novel Rademacher complexity style bound on multicalibration uniform convergence. Beyond these theoretical contributions, we explored our bounds and the overarching concept of multicalibration uniform convergence with experimentation using a two-layer ReLU network, linear SVM, and RBF SVM. Our results provide ML practitioners, equipped with ERM learning bounds, a plug-and-play technique that yields sample complexity bounds for multicalibration uniform convergence. 

In future work, we will explore the empirical behavior of multicalibration error convergence in larger datasets and machine learning applications such as voice recognition and image recognition. As machine learning fairness is intrinsically a human-centered problem, we will focus on how the existing theory for fair classifiers matches the performance of such classifiers in these more human-focused settings. In a more theoretical vein, sharp sample complexity lower bounds for multicalibration error also present an interesting research direction.

\newpage
\bibliographystyle{unsrt}
\bibliography{main}
\appendix

   \section{Additional Figures}\label{sec:AdditionalFigures}
   
   In this section we discuss the preparation of datasets for plotting the race attribute.
   
   \subsubsection*{Adult dataset}

The Adult dataset from the UCI machine learning repository comprises 14 attributes and 48842 records.  Entries in the dataset are extracted from the 1994 Census database. The label indicates if the income of a person exceeds \$50K. With respect to race, there are 470 entries with American Indian and Eskimo as the race attribute, 1519 entries with Asian and Pacific-Islander as the race attribute, 4685 entries with Black as the race attribute, 41762 entries with White as the race attribute, and 406 with Other as the race attribute.

For the race attribute, when implementing the procedure previously defined in \cref{sec:evaluation},  the two population groups, $\group_1$ and $\group_2$ correspond to Black and White, respectively. We use $\placeholderSet_{\mathrm{White}} = \placeholderSet_{\mathrm{Black}} = \left\{20,40,80,160,320,640,1280,2560\right\}$. When plotting the race attribute, only considered are classifiers trained on between 2500 and 5000 examples.

        % $\vec{\placeholder}_{\mathrm{white}} = \left\{20,40,80,160,320,640,1280,2560\right\}$
        
        For each train-test split,  we train  a Radial Basis Function (RBF) Kernel Support Vector Machine (SVM), Linear SVM,  and a two-layer ReLU network consisting of 1000 hidden units.
        
\begin{figure}[b]
    \centering
        \includegraphics[width=1.0\textwidth]{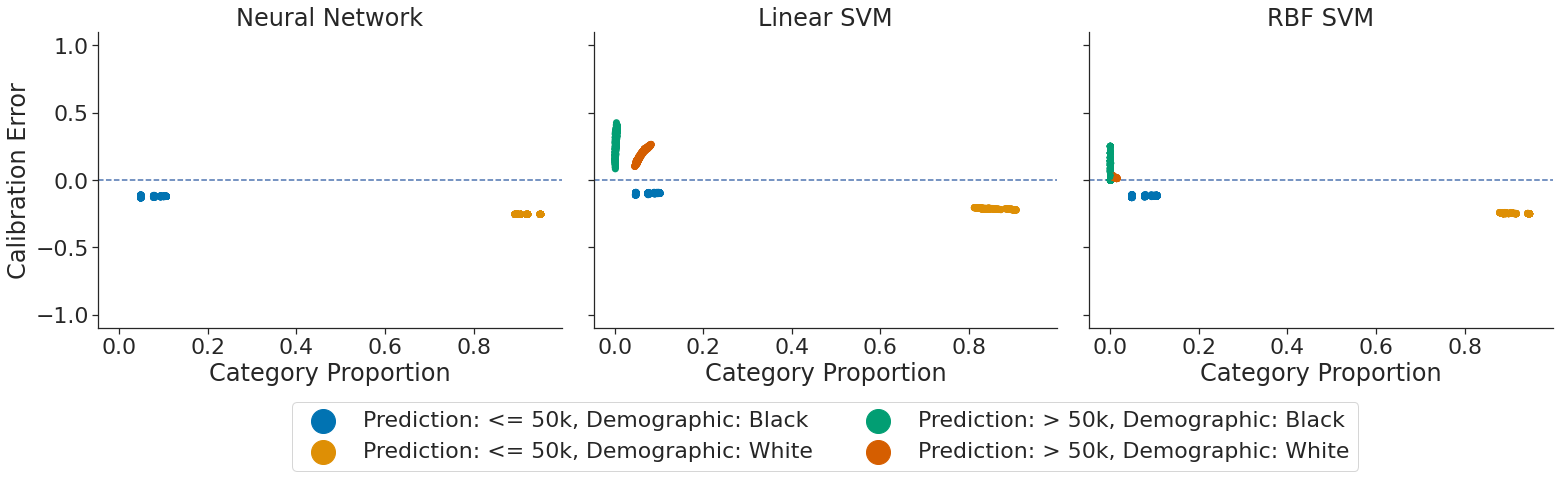}
    \caption{The multicalibration convergence behavior on the UCI Adult dataset, conditioned by race.}
     \label{fig:AdultRace}
\end{figure}

\Cref{fig:AdultRace} depicts the convergence behavior of multicalibration error on the UCI Adult dataset, conditioned by race.
\subsubsection*{COMPAS dataset.}
        
The COMPAS dataset from ProPublica. This dataset contains the features utilized by the COMPAS algorithm in assigning the risk of reoffending for defendants in Broward County, Florida. With respect to race, there are 3175 individuals with Black as the race attribute, 2103 individuals with White as the race attribute, 509 with Hispanic as the race attribute, 31 with Asian as the race attribute, 11 with Native American as the race attribute, and 343 individuals with a race attribute listed as Other. the The dataset comprises roughly 7200 records and 28 features; the label indicates if a defendant re-offends within two years. For the race attribute, the two population groups, $\group_1$ and $\group_2$ correspond to Black and White, respectively. We use $\placeholderSet_{\mathrm{White}} = \placeholderSet_{\mathrm{Black}} = \left\{20,40,80,160,320,640,1280\right\}$.

 For both the race attributes, plots are generated for classifiers trained on between 800 and 1600 examples. For each train-test split,  we train a RBF Kernel SVM, Linear SVM, and a two-layer ReLU network consisting of 1000 hidden units.

\begin{figure}[b]
    \centering
        \includegraphics[width=1.0\textwidth]{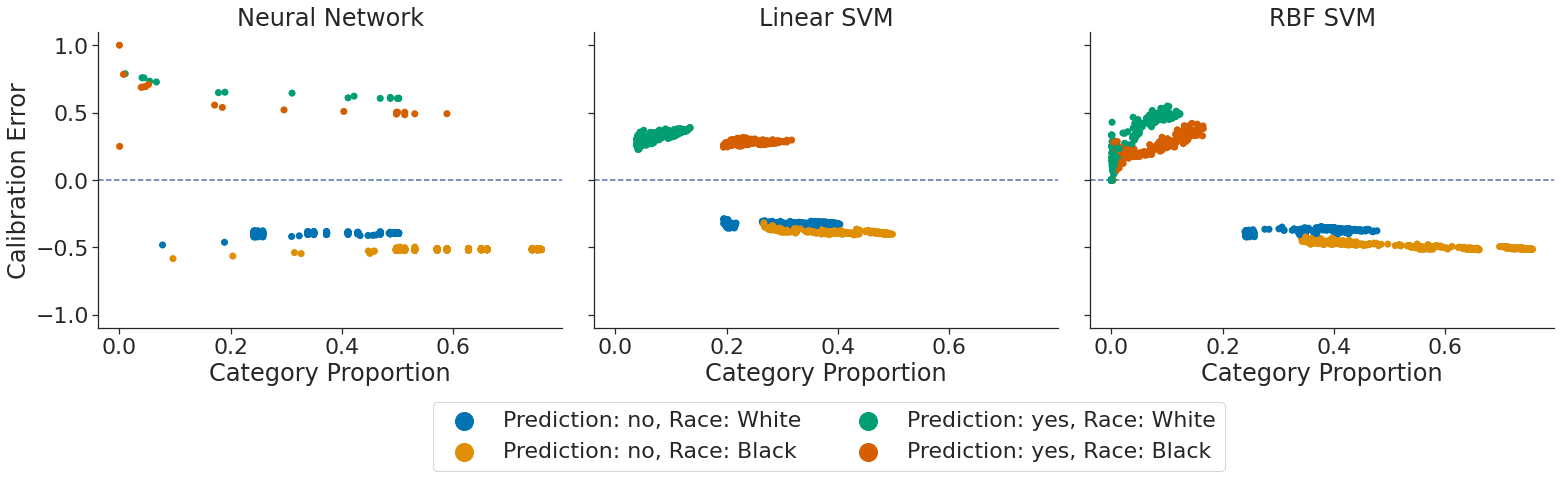}
    \caption{The multicalibration convergence behavior on the COMPAS dataset, conditioned by race.  }
    \label{fig:CompasRace}
\end{figure}

\Cref{fig:CompasRace} depicts the convergence behavior of multicalibration error on the COMPAS dataset, conditioned by race.

\section{Proof of Main Theorem}\label{sec:ProveMainTheorem}

Our proof of \cref{thm:maintheorem} follows the proof technique of theorem 15 from Shabat et al \cite{shabat2020sample}.  The principal difference is that we utilize general sample complexities for ERM learning instead of devising an VC dimension specific bound. Let us begin by assuming a group-wise sample complexity exists for each group, that is:

A function $\sampleComplexity_{\group,\hypothesisClass} : (0,1)^2 \to \naturalNumbers$ such that for every $\errorTolerance,\errorConfidence \in (0,1)$ and every probability distribution in group $\group$, $\sampleDist_{\group}$, if $\numSamples_\group \geq \sampleComplexity_{\group,\hypothesisClass}(\errorTolerance,\errorConfidence)$, we have

\begin{equation}
\Pr\left[\abs{\risk_{\trainingData_{\group}}(\hyp) - \risk_{\sampleDist_{\group}}(\hyp)} \leq \errorTolerance \right] \geq 1-\errorConfidence
\end{equation}

Indeed, this formulation allows use any sort of concentration bound, including those based upon Rademacher complexity and VC dimension.  Once we have a general concentration bound on calibration error for a demographic group, we leverage the frequency of each group to yield an overall bound for multicalibration uniform convergence .

To make our proof, we will also need the following lemmas:

\begin{lemma}[Lemma 26 from Shabat et al. \cite{shabat2020sample}]\label{lem:shabat26}
   Let $p_1,p_2,\tilde{p}_1,\tilde{p}_2,\epsilon,\psi \in [0,1]$ such that $p_1,\psi \leq p_2$ and $\left|p_1-\tilde{p}_1\right|,\left|p_2-\tilde{p}_2\right| \leq \psi\epsilon/3$.  Then,

   \begin{equation}
      \left|\frac{p_1}{p_2} - \frac{\tilde{p}_1}{\tilde{p}_2}\right| \leq \epsilon
   \end{equation}

\end{lemma}

\begin{lemma}[Lemma 27 from Shabat et al. \cite{shabat2020sample}]\label{lem:shabat27}
  Let $\gamma \in (0,1)$ and let $\Gamma_\gamma = \left\{\group \in \Gamma \given \Pr_{\sample \drawnFrom \sampleDist}\left[\sample \in \group\right] \geq \gamma\right\}$ be the collection of subpopulations from $\Gamma$ that has probability at least $\gamma$ according to $\sampleDist$.  Let $\errorConfidence \in (0,1)$ and let $\trainingData$ be a random sample of $\numSamples$ drawn i.i.d. from $\sampleDist$.  Then with probability at least $1-\errorConfidence$, if $\numSamples \geq \frac{8\log\left(\frac{\left|\Gamma\right|}{\errorConfidence}\right)}{\gamma}$,

  \begin{equation}
      \forall \group \in \groups \hiderel{:} \left|\trainingData \intersection \group\right| \hiderel{>} \frac{\gamma \numSamples}{2}
  \end{equation}

\end{lemma}

The proof of \cref{thm:maintheorem} follows:

\begin{proof} Let $\trainingData \definedAs \left\{(\sample_i,\labelElement_i)\right\}_{i=1}^{\numSamples}$ be a sample of $\numSamples$ labeled examples drawn i.i.d. from $\sampleDist$, and let $\trainingData_{\group} \definedAs \left\{(\sample,\labelElement) \in \trainingData : \sample \in {\group} \right\}$ be the samples in $\trainingData$ belonging to group ${\group}$. Let $\hypothesisClass$ denote a fixed hypothesis class and assume $\sampleComplexity_{\group,\hypothesisClass}(\errorTolerance,\errorConfidence)$ exists.  Recall that $\gamma$ is the lowest frequency amongst groups $\group \in \groups$ present in the training data $\trainingData$:

   Let us assume $\numSamples \geq \frac{8\log\left(\frac{2\left|\groups\right|}{\errorConfidence}\right)}{\gamma}$.  This assumption allows us to invoke \cref{lem:shabat27}, which tells that with probability $1-\frac{\errorConfidence}{2}$, for every $\group \in \groups$:
   \begin{equation}\label{eq:minSampleSize}
      \left|\trainingData_{\group}\right| \geq \frac{\gamma \numSamples}{2}
   \end{equation}

   Next, we show that, with high probability, having a large number of examples in $\trainingData_\group$ yields an accurate estimate of the calibration error for an interesting category $(\group,\prediction)$.  For this purpose, we define
   \begin{align}
      \errorTolerance' &\definedAs \frac{\psi\errorTolerance}{3} & \errorConfidence' &\definedAs \frac{\errorConfidence}{4\left|\groups\right|\left|\labelSpace\right|}
   \end{align}

   Given the existence of $\sampleComplexity_{\group,\hypothesisClass}(\errorTolerance,\errorConfidence)$, we know for any $\predictionValue \in \labelSpace$ and any ${\group} \in \groups_{\gamma}$, with probability at least $1-\errorConfidence'$, for a random sample of $\numSamples_1$ examples from ${\group}$ where   
   \begin{align}
      \numSamples_1 &\hiderel{=} \sampleComplexity_{\group,\hypothesisClass}\left(\errorTolerance',\errorConfidence'\right) \\
      &=  \sampleComplexity_{\group,\hypothesisClass}\left(\frac{\psi\errorTolerance}{3}, \frac{\errorConfidence}{4\left|\groups\right|\left|\labelSpace\right|}\right)
\end{align}
   then we have

\begin{equation}
   \forall \hyp \hiderel{\in} \hypothesisClass_{\kernel} \hiderel{:} \left|\frac{1}{\numSamples_{\group}}\sum_{\sample' \hiderel{\in} \trainingData_{\group}}\indicatorFunction\left[\hyp(\sample') \hiderel{=} \predictionValue\right]  \hiderel{-}\Pr_{\sample \drawnFrom \sampleDist_{\group}}\left[\hyp(\sample)\hiderel{=} \predictionValue\right]\right| \hiderel{\leq} \frac{\psi\errorTolerance}{3}
\end{equation}
 Given the existence of $\sampleComplexity_{\group,\hypothesisClass}(\errorTolerance,\errorConfidence)$ we also know that for any $\predictionValue \in \labelSpace$ and any group ${\group} \in \groups_\gamma$, with probability at least $1-\errorConfidence'$, for a random sample of $\numSamples_2$ labeled examples from ${\group} \times \labelSpace$, where

\begin{align}
      \numSamples_2 &\hiderel{=} \sampleComplexity_{\group,\hypothesisClass}\left(\frac{\psi\errorTolerance}{3}, \frac{\errorConfidence}{4\left|\groups\right|\left|\labelSpace\right|}\right)
\end{align}

then we also have 

   \begin{equation}
         \forall \hyp \hiderel{\in} \hypothesisClass_{\kernel} \hiderel{:} \left|\frac{1}{\numSamples_\group}\sum_{(\sample,\labelElement) \in \trainingData_{\group}}\indicatorFunction\left[\hyp(\sample_i) \hiderel{=} \predictionValue,\labelElement\hiderel{=}1\right]  \hiderel{-}\Pr_{\sample \drawnFrom \sampleDist_{\group}}\left[\hyp(\sample)\hiderel{=} \predictionValue,\labelElement\hiderel{=}1\right]\right| \hiderel{\leq} \frac{\psi\errorTolerance}{3}
   \end{equation}
   Set $\numSamples^\marker$ such that upper-bounds $\max\left\{\numSamples_1, \numSamples_2\right\}$
   \begin{equation}
    \numSamples^\marker =  2\sampleComplexity_{\group,\hypothesisClass}\left(\frac{\psi\errorTolerance}{3}, \frac{\errorConfidence}{4\left|\groups\right|\left|\labelSpace\right|}\right)
   \end{equation}
   Then, if for all groups ${\group} \in \groups_\gamma$, $\left|\trainingData_{\group}\right| \geq \numSamples^\marker$, we have with probability at least $1-2\left|\groups\right|\left|\labelSpace\right|\errorConfidence' = 1 - \frac{\errorConfidence}{2}$:
   
   \begin{align} 
        \begin{split} \label{eq:calibrationClosenessA}
       &\forall \hyp \hiderel{\in} \hypothesisClass_{\kernel}, \forall {\group}\hiderel{\in} \groups, \forall \predictionValue \in \labelSpace : \\
       &\left|\frac{1}{\numSamples_{\group}}\sum_{\sample' \in \trainingData_{\group}}\indicatorFunction\left[\hyp(\sample') \hiderel{=} \predictionValue\right]  -\Pr_{\sample \drawnFrom \sampleDist_{\group}}\left[\hyp(\sample)\hiderel{=} \predictionValue\right]\right| \hiderel{\leq} \frac{\psi\errorTolerance}{3}
       \end{split}\\
       \shortintertext{and we also have}
       \begin{split} \label{eq:calibrationClosenessB}
            &\forall \hyp \hiderel{\in} \hypothesisClass_{\kernel}, \forall {\group}\hiderel{\in} \groups, \forall \predictionValue \in \labelSpace :\\
            & \left|\frac{1}{\numSamples_\group}\sum_{(\sample,\labelElement) \in \trainingData_{\group}}\indicatorFunction\left[\hyp(\sample_i) \hiderel{=} \predictionValue,\labelElement\hiderel{=}1\right]  -\Pr_{\sample \drawnFrom \sampleDist_{\group}}\left[\hyp(\sample)\hiderel{=} \predictionValue,\labelElement\hiderel{=}1\right]\right| \hiderel{\leq} \frac{\psi\errorTolerance}{3}
       \end{split}
   \end{align}
   
   Let us now choose the following overall sample size :
   \begin{dmath}
      \numSamples \definedAs   2\sampleComplexity_{\group,\hypothesisClass}\left(\frac{\psi\errorTolerance}{3}, \frac{\errorConfidence}{4\left|\groups\right|\left|\labelSpace\right|}\right)
   \end{dmath}
Recall our use of \cref{lem:shabat27} at the beginning of this proof.  Given a sample of size at least $\numSamples$, we know with probability at least $1-\frac{\errorConfidence}{2}$, for every ${\group} \in \groups_\gamma$. Recall there are at most $\left|\groups \right|$ such population subgroups, and with probability at least $1-\errorConfidence/2$, for every $\group \in \groups_{\gamma}$:
\begin{dmath}
   \left|\trainingData_{\group}\right| \geq \frac{\gamma \numSamples}{2} \hiderel{=}\numSamples^{\marker}
\end{dmath}
   By making the assumptions in \eqref{eq:minSampleSize}, \eqref{eq:calibrationClosenessA},  \eqref{eq:calibrationClosenessB}, and invoking the union bound again, we have, with probability at least $1-\errorConfidence$
   \begin{align}
      \begin{split}
         &{\forall \hyp \hiderel{\in} \hypothesisClass_{\kernel}, \forall {\group}\hiderel{\in} \groups, \forall \predictionValue \in \labelSpace :}\\ &\left|\frac{1}{\numSamples_{\group}}\sum_{\sample' \in \trainingData_{\group}}\indicatorFunction\left[\hyp(\sample') \hiderel{=} \predictionValue\right]  -\Pr_{\sample \drawnFrom \sampleDist_{\group}}\left[\hyp(\sample)\hiderel{=} \predictionValue\right]\right| \hiderel{\leq} \frac{\psi\errorTolerance}{3}
      \end{split} \\
      \shortintertext{and we also have}
      \begin{split}
         &{\forall \hyp \hiderel{\in} \hypothesisClass_{\kernel}, \forall {\group}\hiderel{\in} \groups, \forall \predictionValue \in \labelSpace:}\\   &\left|\frac{1}{\numSamples_\group}\sum_{i=1}^{\numSamples}\indicatorFunction\left[\hyp(\sample_i) \hiderel{=} \predictionValue,\labelElement\hiderel{=}1\right]  -\Pr_{\sample \drawnFrom \sampleDist_{\group}}\left[\hyp(\sample)\hiderel{=} \predictionValue,\labelElement\hiderel{=}1\right]\right| \hiderel{\leq} \frac{\psi\errorTolerance}{3}
      \end{split}
   \end{align}
To conclude the proof, we will invoke \cref{lem:shabat26}.  For notational convenience, we denote the following:
\begin{align}
      p_1(\hyp,{\group},\predictionValue) &\definedAs \Pr\left[\hyp(\sample)\hiderel{=}\predictionValue, \labelElement\hiderel{=}1 \given \sample \hiderel{\in} {\group}\right] \\
      p_2(\hyp,{\group},\predictionValue) &\definedAs \Pr\left[\hyp(\sample)\hiderel{=}\predictionValue, \given \sample \hiderel{\in} {\group}\right] \\
      \tilde{p}_1(\hyp,{\group},\predictionValue) &\definedAs \frac{1}{\left|\trainingData_{\group}\right|}\sum_{(\sample',\labelElement')\hiderel{\in} \trainingData_{\group}}\indicatorFunction\left[\hyp(\sample)\hiderel{=}\predictionValue, \labelElement\hiderel{=}1 \right] \\
      \tilde{p}_2(\hyp,{\group},\predictionValue) &\definedAs \frac{1}{\left|\trainingData_{\group}\right|}\sum_{\sample'\hiderel{\in} \trainingData_{\group}} \indicatorFunction\left[\hyp(\sample)\hiderel{=}\predictionValue\right]
\end{align}

Then, upon invoking \cref{lem:shabat26}, we have for all $\hyp \in \hypothesisClass_{\kernel}, {\group} \in \groups_{\gamma}$ and $\predictionValue \in \labelSpace$, if $p_2(\hyp,\group,\predictionValue) \geq \psi$:

\begin{equation}
   \left|\frac{p_1(\hyp,{\group},\predictionValue)}{p_2(\hyp,\group,\predictionValue)} - \frac{\tilde{p}_1(\hyp,{\group},\predictionValue)}{\tilde{p}_2(\hyp,{\group},\predictionValue)}\right| \leq \epsilon
\end{equation}

Thus, with probability at least $1-\errorConfidence$, we have

\begin{align}
   {\forall \hyp \hiderel{\in} \hypothesisClass_{\kernel}, \forall {\group}\hiderel{\in} \groups, \forall \predictionValue \in \labelSpace:} & \quad \Pr\left[\sample \in {\group}\right] \hiderel{\geq} \gamma \text{ and }  \Pr\left[\hyp(\sample)=\predictionValue \given \sample \in {\group} \right] \hiderel{\geq} \psi
\end{align}

which implies 
   $\left|\calibrationError(\hyp,{\group},\predictionValue) \hiderel{-} \hat{\calibrationError}(\hyp,{\group},\predictionValue,\trainingData) \right| \hiderel{\leq} \errorTolerance$
which completes the proof.

\end{proof}

\section{Two-sided Rademacher Complexity Bounds}\label{sec:rad_appendix}

In this section, we discuss the motivation behind the one-sided Rademacher complexity based generalization bounds of Shalev-Shwartz and Ben-David \cite{shalev2014understanding}, and we show how to transform these bounds into two-sided generalization bounds.  Notation and proof techniques in this section largely follow that of Shalev-Shwartz and Ben-David.  Furthermore, the notation is also largely consistent with notation of Shabat et al \cite{shabat2020sample}.  Using the Rademacher random variable, Bartlett et al. \cite{bartlett2002rademacher} were able to develop a notion of expressiveness for a given function class.  This measure of expressiveness is known as the Rademacher Complexity:

\begin{definition}[Rademacher Complexity]\label{def:rademacherComplexity}
The Rademacher complexity of a function class $\mathcal{H}$ with respect to a sample $\trainingData$ is
\begin{equation}
\rademacherComplexity\left(\mathcal{H} \circ \trainingData \right) \definedAs \frac{1}{\numSamples}\E_{\vec{\rademacherRV}\drawnFrom\left\{\pm 1\right\}}\left[\sup_{f \in \mathcal{H}}\sum_{i=1}^{\numSamples}\rademacherRV_i f(\vec{\placeholder}_i)\right]
\end{equation}
where sample $\trainingData \definedAs \left\{\vec{\placeholder}_1,\dotsc,\vec{\placeholder}_\numSamples \right\}$ is a set of of $\numSamples$ i.i.d. samples drawn from distribution $\sampleDist$.
and $\mathcal{H} \circ \trainingData$ denotes the set of all possible evaluations a function $f \in \mathcal{H}$ can achieve on a sample $\trainingData$, i.e.
\begin{equation}
    \mathcal{H} \circ \trainingData \definedAs \left\{(f(\vec{\placeholder}_1),\dotsc,f(\vec{\placeholder}_\numSamples)) : f \in \mathcal{H}\right\}
\end{equation}
\end{definition}

The Rademacher complexity (\cref{def:rademacherComplexity}) is a useful notion of function class expressiveness because it can be used to bound the \emph{representativeness} of a sample $\trainingData$ with respect to $\loss \circ \mathcal{H}$ as the largest gap between the true risk of a function $\hyp$ and its empirical risk.

\begin{definition}
    The \emph{representativeness} of $\trainingData$, a sample drawn from distribution $\sampleDist$, with respect to $\mathcal{H}$, denoted $\Representativeness_{\sampleDist}(\loss \circ \mathcal{H},\trainingData)$, is the largest gap between the true risk of a hypothesis $\hyp$ in hypothesis class $\hypothesisClass$ and its empirical risk.
\begin{equation}
    \Representativeness_{\sampleDist}(\loss \circ \mathcal{H},\trainingData) \definedAs \sup_{\hyp \in \mathcal{H}} \left(L_{\sampleDist}(\hyp) - L_{\trainingData}(\hyp)\right)
\end{equation}
\end{definition}

The following lemma, from Shalev-Shwartz and Ben-David  describes the relationship between expected representativeness and expected Rademacher complexity:

\begin{lemma}[Lemma 26.2 from Shalev-Shwartz and Ben-David \cite{shalev2014understanding}]\label{lem:representativeness}
    \begin{equation}
\E_{\trainingData \drawnFrom \sampleDist} \Representativeness_{\sampleDist}(\loss \circ \mathcal{H},\trainingData) \leq 2 \E_{\trainingData \drawnFrom \sampleDist} \rademacherComplexity_{\trainingData}\left(\loss\circ\mathcal{H} \right) 
    \end{equation}
\end{lemma}

Inspired by this bound, it is possible to construct generalization bounds based on empirical Rademacher complexity:

\begin{theorem}[Theorem 26.5 from Shalev-Shwartz and Ben-David \cite{shalev2014understanding}]\label{thm:shalevRademacher}
    Let $\trainingData \drawnFrom \sampleDist^{\numSamples}$ be an i.i.d. sample of size $\numSamples$. Assume that for all $(\sample,\labelElement) \in \trainingData$ and $\hyp \in \hypothesisClass$, we have that $\left|\loss(\hyp(\sample),\labelElement)\right| \leq c$. Then
    with probability of at least $1-\errorConfidence$, for all $\hyp \in \hypothesisClass$:
    \begin{align}
        L_\sampleDist(\hyp) - L_{\trainingData}(\hyp) &\leq 2\E_{\trainingData' \drawnFrom \sampleDist^{\numSamples}} \rademacherComplexity_{\trainingData'}(\loss \circ \hypothesisClass) + c\sqrt{\frac{2\ln\left(2/\errorConfidence\right)}{\numSamples}} \\
        L_\sampleDist(\hyp) - L_{\trainingData}(\hyp) &\leq 2\rademacherComplexity_{\trainingData}(\loss \circ \hypothesisClass) + 4c\sqrt{\frac{2\ln\left(4/\errorConfidence\right)}{\numSamples}}
\end{align}
\end{theorem}

Note that \cref{thm:shalevRademacher} contains only one-sided bounds on $L_\sampleDist(\hyp) - L_{\trainingData}(\hyp)$, yet we would like two-sided bounds, i.e. bounds on the probability $\left|L_{\sampleDist}(\hyp) - L_{\trainingData}(\hyp)\right|$ is small.  Let us now pursue such bounds $\left|L_{\sampleDist}(\hyp)-L_{\trainingData}(\hyp)\right|$ using the Rademacher complexity.  Our proof follows the style of the proof of lemma 26.2 from Shalev-Shwartz and Ben-David \cite{shalev2014understanding}.  

\begin{lemma}[Two-sided version of \cref{lem:representativeness}]\label{lem:improvedRepresentativeness}
    \begin{equation}
    \E_{\trainingData \drawnFrom \sampleDist^\numSamples} \sup_{\hyp \in \mathcal{H}} \left|L_{\sampleDist}(\hyp) - L_{\trainingData}(\hyp)\right| \leq 2 \E_{\trainingData \drawnFrom \sampleDist^{\numSamples}} \rademacherComplexity_{\trainingData}\left(\loss \circ \mathcal{H} \right) 
    \end{equation}
\end{lemma}
\begin{proof}
Let us now provide a proof of  \cref{lem:improvedRepresentativeness}:

\begin{align}
    \sup_{\hyp \in \mathcal{H}} \left| L_{\sampleDist}(\hyp)-L_{\trainingData}(\hyp) \right| &= \sup_{\hyp \in \mathcal{H}} \E_{\trainingData' \drawnFrom \sampleDist^{\numSamples}}\left|L_{\trainingData'}(\hyp)-L_{\trainingData}(\hyp)\right| \\
    \intertext{Because expectation of supremum is larger than supremum of expectation, we have:}
        \sup_{\hyp \in \mathcal{H}} \left| L_{\sampleDist}(\hyp)-L_{\trainingData}(\hyp) \right| &\leq \E_{\trainingData' \drawnFrom \sampleDist^{\numSamples}} \sup_{\hyp \in \mathcal{H}} \left|L_{\trainingData'}(\hyp)-L_{\trainingData}(\hyp)\right|
    \end{align}
    Taking expectation over $\trainingData$ on both sides, we have
    \begin{align}
        \E_{\trainingData \drawnFrom \sampleDist^{\numSamples}}\left[\sup_{\hyp \in \mathcal{H}} \left| L_{\sampleDist}(\hyp)-L_{\trainingData}(\hyp) \right|\right] &\leq \E_{\trainingData,\trainingData' \drawnFrom \sampleDist^{\numSamples}}\left[\sup_{\hyp \in \mathcal{H}} \left|L_{\trainingData'}(\hyp)-L_{\trainingData}(\hyp)\right|\right] \\
            &=\frac{1}{\numSamples} \E_{\trainingData,\trainingData' \drawnFrom \sampleDist^{\numSamples}}\left[\sup_{\hyp \in \mathcal{H}}  \left| \sum_{i=1}^\numSamples \loss(\hyp(\sample_i'),\labelElement_i') - \loss(\hyp(\sample_i),\labelElement_i)\right|\right]
        \end{align}
        For each $j$, $(\sample_j',\labelElement_j')$ and $(\sample_j,\labelElement_j)$ are i.i.d. variables, which allows us to make the following statement:
        \begin{align}
            \underbrace{\frac{1}{\numSamples}\E_{\trainingData,\trainingData' \drawnFrom \sampleDist^{\numSamples}}\left[\sup_{\hyp \in \mathcal{H}}  \left|\loss(\hyp(\sample_j'),\labelElement_j') - \loss(\hyp(\sample_j),\labelElement_j) +\sum_{i\neq j} \loss(\hyp(\sample_i'),\labelElement_i') - \loss(\hyp(\sample_i),\labelElement_i)\right|\right]}_{A}\\
            = \underbrace{\frac{1}{\numSamples}\E_{\trainingData,\trainingData' \drawnFrom \sampleDist^{\numSamples}}\left[\sup_{\hyp \in \mathcal{H}}  \left|\loss(\hyp(\sample_j),\labelElement_j) - \loss(\hyp(\sample_j'),\labelElement_j') +\sum_{i\neq j} \loss(\hyp(\sample_i'),\labelElement_i') - \loss(\hyp(\sample_i),\labelElement_i)\right|\right]}_{B} 
        \end{align}
        \begin{dgroup*}
        \intertext{Let $\rademacherRV_j$ be a Rademacher random variable (i.e. a draw from the uniform distribution over $\left\{+1,-1\right\}$).  Then we have:}
        \begin{dmath}
            \frac{1}{\numSamples}\E_{\trainingData,\trainingData' \drawnFrom \sampleDist^{\numSamples}}\left[\sup_{\hyp \in \mathcal{H}}  \left|\rademacherRV_j(\loss(\hyp(\sample_j'),\labelElement_j') - \loss(\hyp(\sample_j),\labelElement_j)) +\sum_{i\neq j} \loss(\hyp(\sample_i'),\labelElement_i') - \loss(\hyp(\sample_i),\labelElement_i)\right|\right] = \frac{1}{2}\left(A + B\right)
            \end{dmath}
            \begin{dmath}
            = \frac{1}{\numSamples}\E_{\trainingData,\trainingData' \drawnFrom \sampleDist^{\numSamples}}\left[\sup_{\hyp \in \mathcal{H}}  \left|\loss(\hyp(\sample_i'),\labelElement_i') - \loss(\hyp(\sample_i),\labelElement_i)\right|\right]
            \end{dmath}
        \end{dgroup*}
        \begin{dgroup*}
        \intertext{Let $\vec{\rademacherRV} = \left[\rademacherRV_1,\dotsc,\rademacherRV_{\numSamples}\right]^{\transpose}$ be a vector consisting of $\numSamples$ i.i.d. samples from the Rademacher distribution.  Repeating this procedure for all $j$, we have}
        \begin{dmath}
            \frac{1}{\numSamples}\E_{\trainingData,\trainingData' \drawnFrom \sampleDist^{\numSamples}}\left[\sup_{\hyp \in \mathcal{H}}  \left|\sum_{i=1}^{\numSamples} \loss(\hyp(\sample_i'),\labelElement_i') - \loss(\hyp(\sample_i),\labelElement_i)\right|\right] =  \frac{1}{\numSamples}\E_{\trainingData,\trainingData' \drawnFrom \sampleDist^{\numSamples},\vec{\rademacherRV}}\left[\sup_{\hyp \in \mathcal{H}}  \left|\sum_{i=1}^{\numSamples} \rademacherRV_i(\loss(\hyp(\sample_i'),\labelElement_i') - \loss(\hyp(\sample_i),\labelElement_i)\right|\right] 
            \end{dmath}
            \begin{dmath}
            = \frac{1}{\numSamples}\E_{\trainingData,\trainingData' \drawnFrom \sampleDist^{\numSamples},\vec{\rademacherRV}}\left[\sup_{\hyp \in \mathcal{H}}  \left(\sum_{i=1}^{\numSamples} \rademacherRV_i\loss(\hyp(\sample_i'),\labelElement_i') \right) - \inf_{\hyp \in \mathcal{H}}\left(\sum_{i=1}^{\numSamples}  \rademacherRV_i\loss(\hyp(\sample_i),\labelElement_i) \right)\right] \\
            \end{dmath}
        \intertext{We know the distribution of $\sum_{i=1}^{\numSamples}  \rademacherRV_i\loss(\hyp(\sample_i),\labelElement_i)$ to be symmetric, hence we have}
            \begin{dmath}
            =\frac{1}{\numSamples}\E_{\trainingData,\trainingData' \drawnFrom \sampleDist^{\numSamples},\vec{\rademacherRV}}\left[\sup_{\hyp \in \mathcal{H}}  \left(\sum_{i=1}^{\numSamples} \rademacherRV_i\loss(\hyp(\sample_i'),\labelElement_i') \right) \hiderel{+} \sup_{\hyp \in \mathcal{H}}\left(\sum_{i=1}^{\numSamples}  \rademacherRV_i\loss(\hyp(\sample_i),\labelElement_i) \right)\right] 
            \end{dmath}
            \begin{dmath}
            =\frac{2}{\numSamples}\E_{\trainingData \drawnFrom \sampleDist^{\numSamples},\vec{\rademacherRV}}\left[\sup_{f \in \mathcal{H}}  \left(\sum_{i=1}^{\numSamples} \rademacherRV_i\loss(\hyp(\sample_i),\labelElement_i) \right) \right]
            \end{dmath}
            \begin{dmath}
            =2\E_{\trainingData \drawnFrom \sampleDist^{\numSamples},\vec{\rademacherRV}}\left[\rademacherComplexity_{\trainingData}(\loss \circ \mathcal{H})\right]
            \end{dmath}
\end{dgroup*}
\end{proof}

Upon replacing \cref{lem:representativeness} with \Cref{lem:improvedRepresentativeness}, \cref{thm:ShalevImproved} naturally follows from the proof of \cref{thm:shalevRademacher}.  Hence, we arrive at the following two-sided bound:

\begin{theorem}[Two-sided version of Theorem 26.5 from Shalev-Shwartz and Ben-David \cite{shalev2014understanding}]\label{thm:ShalevImproved}
    Let $\trainingData \drawnFrom \sampleDist^{\numSamples}$ be a sample of size $\numSamples$. Assume that for all $(\sample,\labelElement) \in \trainingData$ and $\hyp \in \hypothesisClass$, we have that $\left|\loss(\hyp(\sample),\labelElement)\right| \leq c$. Then
    with probability of at least $1-\errorConfidence$, for all $\hyp \in \hypothesisClass$:
    \begin{dgroup*}
    \begin{dmath}
        \left|L_\sampleDist(\hyp) - L_{\trainingData}(\hyp)\right| \leq 2\E_{\trainingData' \drawnFrom \sampleDist^{\numSamples}} \rademacherComplexity_{\trainingData'}(\loss \circ \hypothesisClass) + c\sqrt{\frac{2\ln\left(2/\errorConfidence\right)}{\numSamples}}
    \end{dmath}
    \begin{dmath}\label{eq:rademacherEmpiricalTwoSided}
        \left|L_\sampleDist(\hyp) - L_{\trainingData}(\hyp)\right| \leq 2\rademacherComplexity_{\trainingData}(\loss \circ \hypothesisClass \circ \trainingData) + 4c\sqrt{\frac{2\ln\left(4/\errorConfidence\right)}{\numSamples}}
    \end{dmath}
\end{dgroup*}
\end{theorem}

\section{Bounds, Theorems, and Algebra for Section 3}\label{sec:boundsSec3}

\subsection{Sample Complexities for RBF Kernels}
\begin{theorem}[Simplified version of Theorem 2 from Cortes et al. \cite{cortes2010generalization}]\label{thm:cortesRademacher}
   Let $\kernelFunction$ be a kernel function and that $\kernelFunction(\sample,\sample)\leq \scalarBound^2$ for all $\sample \in \sampleSpace$ .  Then for any sample $\trainingData$ of size $\numSamples$, the Rademacher complexity of the hypothesis class $\hypothesisClass_{\kernelFunction}'$ can be bounded as follows
   \begin{dmath}
      \rademacherComplexity_{\trainingData}(\hypothesisClass_{\kernel}) \leq \sqrt{\frac{23e\scalarBound^2}{22\numSamples}}
   \end{dmath}
\end{theorem}

\begin{lemma}[RBF Kernel SVM Sample Complexity Bound]
   Let $\hypothesisClass_{\kernel}$ be an RBF kernel SVM predictor class. Let $\predictionValue \in \labelSpace$ be a prediction value.  Then for any distribution $\sampleDist$ over $\sampleSpace \times \left\{\pm 1\right\}$ and any $\errorTolerance,\errorConfidence \in (0,1)$, if $\trainingData$ is a random sample of at least $\numSamples$ samples where
   
   \begin{equation}
       \numSamples \geq \frac{23 {\scalarBound^2}+ 64\log\left(4/\errorConfidence\right)}{\errorTolerance^2}
   \end{equation} 
   examples drawn i.i.d. according to $\sampleDist$, then with probability at least $1-\errorConfidence$:
  \begin{equation}
      \forall \hyp \hiderel{\in} \hypothesisClass \hiderel{:} \left|\frac{1}{\placeholderCount}\sum_{i=1}^{\placeholderCount}\indicatorFunction\left[\hyp(\sample_i) \hiderel{=} \predictionValue,\labelElement\hiderel{=}\predictionValue\right]  \hiderel{-}\Pr_{\sample \drawnFrom \sampleDist_U}\left[\hyp(\sample)\hiderel{=} \predictionValue,\labelElement\hiderel{=}\predictionValue\right]\right| \hiderel{\leq} \errorTolerance
  \end{equation}
\end{lemma}

\begin{proof}
Fix $\hyp ,\group, \prediction$ and $\trainingData$. We take $\risk_{\trainingData}(\hyp)$ as $\empiricalCalibrationError(\hyp,\group,\prediction,\trainingData)$ and  $\risk_{\sampleDist}(\hyp)$ as $\calibrationError(\hyp,\group,\prediction)$  for convenience,

Invoking \cref{thm:ShalevImproved}, and solving for $\numSamples$:

 \begin{align}
      \abs{\risk_{\sampleDist}(\hyp) - \risk_{\trainingData}(\hyp)} &\hiderel{\leq} 2\rademacherComplexity_{\trainingData}(\empiricalCalibrationError \circ \hypothesisClass_{\kernel}) + 4c\sqrt{\frac{2\log\left(4/\errorConfidence\right)}{\numSamples}}
      \intertext{Because the absolute value of empirical calibration error is bounded by $1$, we set $c=1$, to yield}
         \abs{\risk_{\sampleDist}(\hyp) - \risk_{\trainingData}(\hyp)}  &\leq 23 \frac{\scalarBound^2}{\numSamples} + 64\frac{\log\left(4/\errorConfidence\right)}{\numSamples}
      \intertext{To guarantee this is smaller the right hand side of the inequality is less than $\epsilon$, we need}
         \numSamples &\geq\frac{23 {\scalarBound^2}+ 64\log\left(4/\errorConfidence\right)}{\errorTolerance^2}
   \end{align}
   
   Thus arriving at the sample complexity bound
\end{proof}

\subsection{Sample Complexities for ReLU Networks}

\begin{lemma}[Two layer ReLU Sample Complexity Bound]

Let $\trainingData \definedAs \left\{(\sample_i,\labelElement_i)\right\}_{i=1}^{\numSamples}$ be a sample of $\numSamples$ labeled examples drawn i.i.d. from $\sampleDist$ Hypothesis class $\hypothesisClass_{\neuralNetwork}$ is parametrized by a sequence of matrices $\weightMatrix = (\weightMatrix_1,\dotsc,\weightMatrix_\depth)$, where $\weightMatrix_{i} \in \realNumbers^{\dimension_{i}\times\dimension_{i-1}}$ where $\dimension = \dimension_0$. In the binary classification setting, we have $\dimension_{\depth} = 1$, and in the multi-class setting, we have $\dimension_{\depth} = \abs{\labelSpace}$.  Let $\nonlinearity(\cdot)$ be the ReLU function.  $\nonlinearity(\vec{\placeholder})$ applies the ReLU to each coordinate in $\vec{\placeholder}$. Such networks are of the form
\begin{equation*}
    \weightMatrix_{\depth}\nonlinearity\left(\weightMatrix_{\depth-1}\nonlinearity\left(\cdots\nonlinearity\left(\weightMatrix_{1}\sample\right)\cdots\right)\right)
\end{equation*}

The following is a Rademacher complexity bound on ReLU networks.  The bound is originally attributed to Bartlett et al.\cite{bartlett2017spectrally}, but we find the manner it is expressed in Theorem 5 from Yin et al. \cite{yin2019rademacher} to be more amenable to our results.
\begin{theorem}[Rademacher Complexity bound for ReLU Networks, \cite{bartlett2017spectrally,yin2019rademacher}]
Consider the neural network hypothesis class
\begin{equation}
    \hypothesisClass_{\neuralNetwork} = \left\{\hyp_{\weightMatrix}(\sample) : \weightMatrix : \left(\weightMatrix_1,\dotsc,\weightMatrix_\depth\right), \norm{\weightMatrix_i}_2 \leq \placeholder_i, \norm{\weightMatrix_i^{\transpose}}_{2,1} \leq \placeholder'_{i}, i \in \left\{1,\dotsc,\depth\right\}\right\}
\end{equation}
Then we have 
\begin{equation*}
    \rademacherComplexity_{\trainingData}(\hypothesisClass_{\neuralNetwork}) \leq \frac{4}{\numSamples^{\sfrac{3}{2}}} + \frac{26\log(\numSamples)\log(2\dimension_{\max})}{\numSamples}\norm{\vec{X}}_{\Frobenius}\left(\prod_{i=1}^{\depth}\placeholder_i\right)\left(\sum_{j=1}^{\depth}\left(\frac{\placeholder_j'}{\placeholder_j}\right)^{\sfrac{2}{3}}\right)^{\sfrac{3}{2}}
\end{equation*}
where $\vec{X}$ are the unlabeled training data $\trainingData$, i.e. $\vec{X} = \left[\sample_1,\dotsc,\sample_N\right]$
\end{theorem}

   Let $\hypothesisClass_{\neuralNetwork}$ be a ReLU neural network predictor class defined above. Let $\predictionValue \in \labelSpace$ be a prediction value and assume 
   
   \begin{equation*}1 \leq\log\left(2\dimension_{\max}\right)\vec{X}_{\Frobenius}\left(\prod_{i=1}^{\depth}\placeholder_i\right)\left(\sum_{j=1}^{\depth}\left(\frac{\placeholder_j'}{\placeholder_j}\right)^{\sfrac{2}{3}}\right)^{\sfrac{3}{2}} \end{equation*}  
   
 Then for any distribution $\sampleDist$ over $\sampleSpace \times \left\{\pm 1\right\}$ and any $\errorTolerance,\errorConfidence \in (0,1)$, if $\trainingData$ is a random sample of at least $\numSamples$ samples where
   
   \begin{equation*}
       \numSamples \geq \frac{7200\log^2\left(2\dimension_{\max}\right)\norm{\vec{X}}_{\Frobenius}^2\left(\prod_{i=1}^{\depth}\placeholder_i\right)^2\left(\sum_{j=1}^{\depth}\left(\frac{\placeholder_j'}{\placeholder_j}\right)^{\sfrac{2}{3}}\right)^{3} + 64\log\left(\sfrac{4}{\errorConfidence}\right)}{\errorTolerance^2}
   \end{equation*} 
   examples drawn i.i.d. according to $\sampleDist$, then with probability at least $1-\errorConfidence$:
  \begin{equation}
      \forall \hyp \hiderel{\in} \hypothesisClass \hiderel{:} \left|\frac{1}{\placeholderCount}\sum_{i=1}^{\placeholderCount}\indicatorFunction\left[\hyp(\sample_i) \hiderel{=} \predictionValue,\labelElement\hiderel{=}\predictionValue\right]  \hiderel{-}\Pr_{\sample \drawnFrom \sampleDist_U}\left[\hyp(\sample)\hiderel{=} \predictionValue,\labelElement\hiderel{=}\predictionValue\right]\right| \hiderel{\leq} \errorTolerance
  \end{equation}
\end{lemma}

\begin{proof} Our proof strategy is to invoke a two-sided Rademacher complexity bound (\cref{thm:ShalevImproved}) to yield a sample complexity bound:
Fix $\hyp ,\group, \prediction$ and $\trainingData$. We take $\risk_{\trainingData}(\hyp)$ as $\empiricalCalibrationError(\hyp,\group,\prediction,\trainingData)$ and  $\risk_{\sampleDist}(\hyp)$ as $\calibrationError(\hyp,\group,\prediction)$ 
     \begin{align}
          \abs{\risk_{\sampleDist}(\hyp) - \risk_{\trainingData}(\hyp)} &\hiderel{\leq} 2\rademacherComplexity_{\trainingData}(\empiricalCalibrationError \circ \hypothesisClass_{\kernel}) + 4c\sqrt{\frac{2\log\left(4/\errorConfidence\right)}{\numSamples}}
      \intertext{Because the absolute value of empirical calibration error is bounded by $1$, we set $c=1$ and substitute the Rademacher complexity of ReLU networks, to yield}
           \abs{\risk_{\sampleDist}(\hyp) - \risk_{\trainingData}(\hyp)}  &\hiderel{\leq} \frac{8}{\numSamples^{\sfrac{3}{2}}} + \frac{52 \log(\numSamples)\log(2\dimension_{\max})}{\numSamples}\norm{\vec{X}}_{\Frobenius}\hiderel{\cdot}\left(\prod_{i=1}^{\depth}\placeholder_i\right)\left(\sum_{j=1}^{\depth}\left(\frac{\placeholder_j'}{\placeholder_j}\right)^{\sfrac{2}{3}}\right)^{\sfrac{3}{2}} \hiderel{+} 4\sqrt{\frac{2\log\left(\sfrac{4}{\errorTolerance}\right)}{\numSamples}} \\
         &\leq \frac{8}{\numSamples^{\sfrac{3}{2}}} + \frac{52 \log(\numSamples)\log(2\dimension_{\max})}{\numSamples^{\sfrac{1}{2}}} \norm{\vec{X}}_{\Frobenius}\left(\prod_{i=1}^{\depth}\placeholder_i\right)\left(\sum_{j=1}^{\depth}\left(\frac{\placeholder_j'}{\placeholder_j}\right)^{\sfrac{2}{3}}\right)^{\sfrac{3}{2}} + 4\sqrt{\frac{2\log\left(\sfrac{4}{\errorTolerance}\right)}{\numSamples}} \\
      \intertext{Let us now assume that }
        1 &\leq\log\left(2\dimension_{\max}\right)\vec{X}_{\Frobenius}\left(\prod_{i=1}^{\depth}\placeholder_i\right)\left(\sum_{j=1}^{\depth}\left(\frac{\placeholder_j'}{\placeholder_j}\right)^{\sfrac{2}{3}}\right)^{\sfrac{3}{2}}
    \intertext{Which gives us}
           \abs{\risk_{\sampleDist}(\hyp) - \risk_{\trainingData}(\hyp)} &\leq  \frac{60 \log(2\dimension_{\max})}{\numSamples^{\sfrac{1}{2}}} \norm{\vec{X}}_{\Frobenius}\left(\prod_{i=1}^{\depth}\placeholder_i\right)\left(\sum_{j=1}^{\depth}\left(\frac{\placeholder_j'}{\placeholder_j}\right)^{\sfrac{2}{3}}\right)^{\sfrac{3}{2}} + 4\sqrt{\frac{2\log\left(\sfrac{4}{\errorTolerance}\right)}{\numSamples}} \\
      \intertext{To guarantee this is smaller the right hand side of the inequality is less than $\epsilon$, we need}
 \numSamples &\geq \frac{7200\log^2\left(2\dimension_{\max}\right)\norm{\vec{X}}_{\Frobenius}^2\left(\prod_{i=1}^{\depth}\placeholder_i\right)^2\left(\sum_{j=1}^{\depth}\left(\frac{\placeholder_j'}{\placeholder_j}\right)^{\sfrac{2}{3}}\right)^{3} + 64\log\left(\sfrac{4}{\errorConfidence}\right)}{\errorTolerance^2} 
\end{align}
   Thus arriving at the sample complexity bound

\end{proof}

 \subsection{VC Dimension Definition}

\begin{definition}[VC dimension]\label{def:VCdimension} Let $\hypothesisClass \subseteq \left\{0, 1\right\}^{\sampleSpace}$ be a binary hypothesis class.  A set $\placeholderSet = \left\{\sample_1,\dotsc,\sample_\numSamples\right\} \subseteq \sampleSpace$ shatters hypothesis class $\hypothesisClass$ if 
    \begin{dmath}
        \left|\left\{\hyp(\sample_1),\dotsc,\hyp(\sample_{\numSamples})\right\} : \hyp \in \hypothesisClass\right| = 2^\numSamples
    \end{dmath}

The VC dimension of $\hypothesisClass$ is the maximal size of $\placeholderSet$ which shatters $\hypothesisClass$.
\end{definition}

\end{document}